\documentclass{article} 
\usepackage{iclr2027_conference,times}

\usepackage{amsmath,amsfonts,bm}

\def\eqref#1{equation~\ref{#1}}

\def\1{\bm{1}}

\DeclareMathAlphabet{\mathsfit}{\encodingdefault}{\sfdefault}{m}{sl}
\SetMathAlphabet{\mathsfit}{bold}{\encodingdefault}{\sfdefault}{bx}{n}

\usepackage{hyperref}
\usepackage{url}
\usepackage[pdftex]{graphicx}

\usepackage{booktabs}
\usepackage{tabularx}
\usepackage{array}
\usepackage{ragged2e}
\usepackage[table]{xcolor}

\usepackage{adjustbox}
\usepackage{multirow}

\newcolumntype{L}[1]{%
  >{\RaggedRight\arraybackslash\hspace{0pt}}p{#1}}
\newcolumntype{C}[1]{%
  >{\Centering\arraybackslash}p{#1}}
\newcolumntype{Y}{%
  >{\RaggedRight\arraybackslash\hspace{0pt}}X}

\newcommand{\name}{EarthAlign}
\newcommand{\numall}{553,738}
\newcommand{\numtrain}{490,395}
\newcommand{\numtestretrieval}{63,343}
\newcommand{\numtestclassification}{48,645}
\newcommand{\numtestwind}{7,353}
\newcommand{\numtestpres}{8,114}
\newcommand{\numnegsampling}{64,553}
\newcommand{\numexpertrecord}{489,185}
\newcommand{\numnpkeywords}{93}

\usepackage{subcaption}
\usepackage{enumitem}
\usepackage[most]{tcolorbox}
\newtcolorbox{promptbox}[1]{
    enhanced,
    breakable,
    colback=gray!5,
    colframe=gray!60,
    colbacktitle=gray!60,
    coltitle=white,
    fonttitle=\bfseries,
    title=#1,
    boxrule=0.6pt,
    arc=2mm,
    left=4mm,
    right=4mm,
    top=2mm,
    bottom=2mm
}

\definecolor{TPIMacro}{HTML}{EAF3F8}
\definecolor{LPMacro}{HTML}{FDF0E4}
\definecolor{newtext}{RGB}{180, 140, 0}

\usepackage{wrapfig}
\usepackage{xurl}

\usepackage{fontawesome5}

\title{\name{}: Learning Language-Aligned Representations of Earth System Data}

\author{\textbf{Fadillah Adamsyah Maani}$^{1}$ \quad
\textbf{Numan Saeed}$^{1}$ \quad
\textbf{Jiaqi Ma}$^{1,*}$ \quad
\textbf{Dat Tien Nguyen}$^{1}$ \\
\textbf{Youssef Wehbe}$^{2}$ \quad
\textbf{Duncan Watson-Parris}$^{3}$ \quad
\textbf{Salman Khan}$^{1,2}$ \\[0.5em]
$^{1}$Mohamed bin Zayed University of Artificial Intelligence \quad
$^{2}$ADIA Lab \\
$^{3}$University of California San Diego \\[0.3em]
$^{*}$Corresponding authors:
\texttt{jiaqi.ma@mbzuai.ac.ae}
\\[0.7em]
\href{https://mbzuai-oryx.github.io/EarthAlign/}{\faGlobe\ \textbf{Website}}
\qquad
\href{https://github.com/mbzuai-oryx/EarthAlign}{\faGithub\ \textbf{Code}}
\qquad
\href{https://huggingface.co/datasets/MBZUAI/EarthAlign}{\faDatabase\ \textbf{Dataset \& Weights}}
}

\iclrfinalcopy

\begin{document}

\maketitle

{
\begin{abstract}
Interpreting Earth system conditions is central to understanding environmental risks and informing decisions across weather- and climate-sensitive sectors. Existing multimodal weather systems primarily target task-specific prediction, generation, or question answering, while general-purpose alignment between Earth system data and language remains underexplored due to the limited availability of diverse language supervision grounded in meteorological conditions. To fill this gap, we introduce \name{}, a large-scale dataset with broad geographic coverage spanning 2000--2024 that combines 10 expert-curated natural event and disaster data sources with climatology-derived normal-weather samples. Each sample includes temporal and geospatial information and four grounded caption styles aligned with 63-variable ERA5 meteorological data. Building on \name{}, we introduce MeteoLIP, a dual-encoder framework for learning aligned meteorology-language representations. MeteoLIP incorporates MetTok, which uses a geolocation- and time-conditioned query to aggregate variable-specific patch embeddings into a compact sequence of meteorology tokens. To comprehensively assess Earth system multimodal representations, we introduce three downstream tasks: cross-modal retrieval, weather-condition classification, and physical variable estimation. Our experiments show that MeteoLIP improves upon the corresponding CLIP and SigLIP baselines across these tasks. Further analyses highlight the importance of meteorological input, temporal and geospatial conditioning, and multi-style supervision for meteorology-language alignment. Thus, \name{} and MeteoLIP provide a foundation for learning and evaluating language-aligned representations of Earth system data. Our code, dataset, and weights are publicly available at \href{https://github.com/mbzuai-oryx/EarthAlign}{https://github.com/mbzuai-oryx/EarthAlign}.
\end{abstract}
}

\begin{figure}[t]
    \begin{center}
    \includegraphics[width=0.9\textwidth]{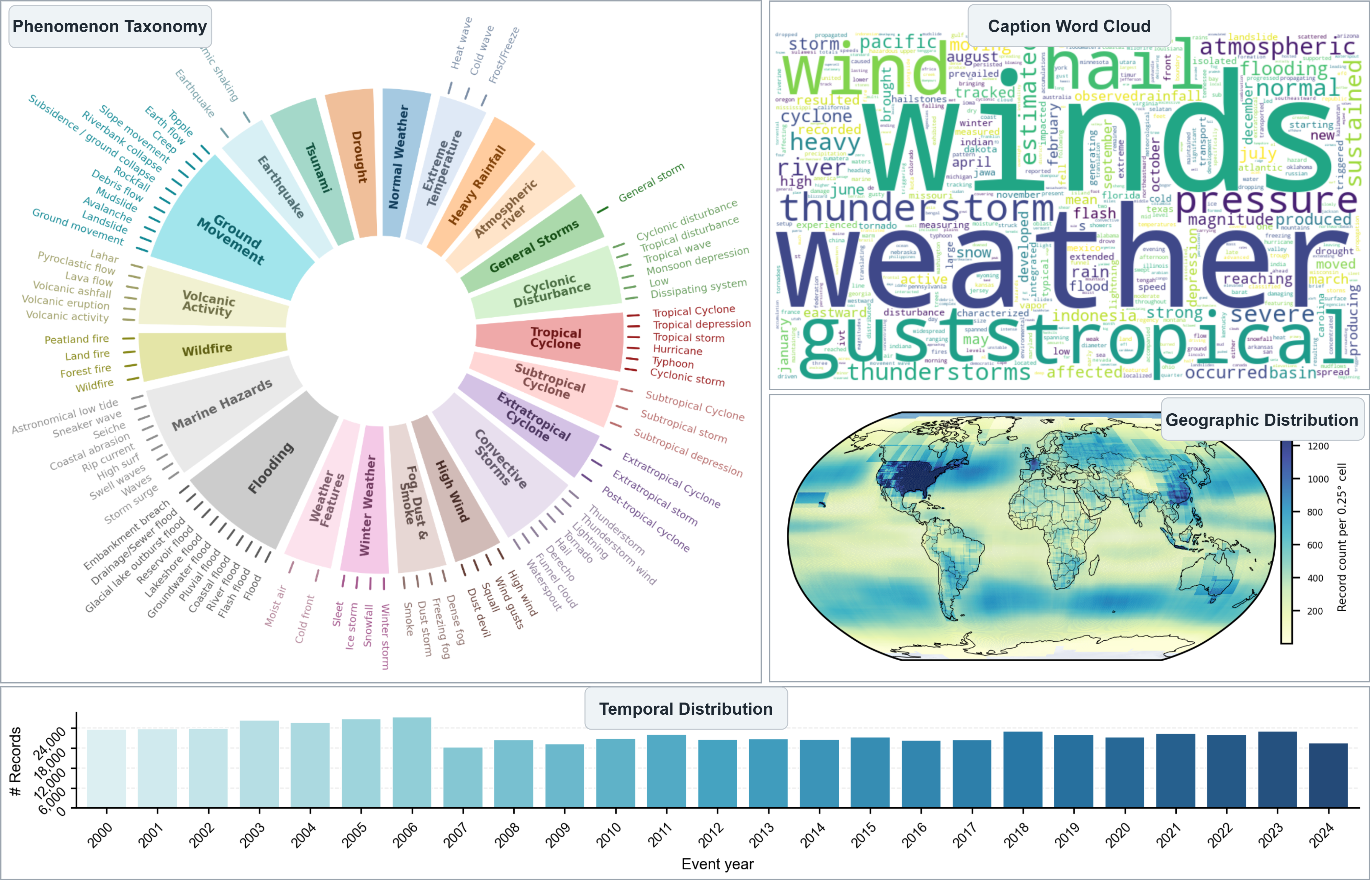}
    \end{center}
    \caption{\textbf{\name{} Overview.}
    \name{} is a vision-language alignment dataset for the Earth system, comprising \numall{} samples that span 2000--2024, with a broad taxonomy of natural phenomena, broad geographic coverage, and diverse caption content to enable multimodal alignment.}
    \label{fig:data_distribution_and_statistics}
\end{figure}

{
\section{Introduction}
Earth system forecasts and analyses play a critical role in supporting decision-making across sectors such as agriculture, water, energy, and disaster management~\citep{brasseur2016climate,white2017s2s}. Recent data-driven weather models and Earth system foundation models, including FourCastNet~\citep{fourcastnet}, Pangu-Weather~\citep{panguweather}, ClimaX~\citep{climax}, and Aurora~\citep{aurorafm}, have achieved strong performance in forecasting atmospheric and broader Earth system states. Their representations are primarily learned through objectives defined over physical variables, such as forecasting future atmospheric states~\citep{climax,aurorafm} and reconstructing masked meteorological fields~\citep{atmorep,prithviwxc}. While these objectives effectively model relationships within numerical Earth system data, they do not explicitly align the learned representations with natural-language descriptions of weather conditions and natural phenomena.

Natural language provides a valuable source of supervision that can align multivariate meteorological data with descriptions of weather conditions and natural phenomena. In natural-image domains, CLIP~\citep{clip} has demonstrated that language supervision can support shared representations for cross-modal retrieval and text-based classification, and such visual representations have subsequently become key components of multimodal models across different domains~\citep{llava,geopixel}. Despite this progress, general-purpose alignment between Earth system data and language remains underexplored, partly due to the lack of diverse language supervision grounded in meteorological conditions and natural phenomena. Existing multimodal datasets such as WeatherQA~\citep{weatherqa}, CLLMate~\citep{cllmate}, and MeteorPred~\citep{meteorpred} have enabled progress in severe-weather question answering, event prediction, and warning generation, but are primarily designed around task-specific prediction, generation, or question answering. Learning robust meteorology-language representations requires rich language supervision covering diverse weather conditions and natural phenomena, with descriptions from different perspectives and temporal and geographic context to associate them with the corresponding Earth system states.

To address this, we construct \name{} (Figure~\ref{fig:data_distribution_and_statistics}), which brings heterogeneous natural event, disaster, and normal-weather records into an Earth system--language setting. \name{} comprises \numall{} samples spanning 2000--2024, with broad geographic coverage and diverse natural phenomena. We compile natural event and disaster records from governmental, institutional, and published sources, and complement them with normal-weather samples identified relative to local climatology. For each record, we generate four metadata-grounded caption styles to provide different semantic perspectives on the underlying Earth system condition. Each record is also paired with a 63-variable ERA5~\citep{era5paper} snapshot at the corresponding time and spatial extent to enable meteorology-language alignment. To learn semantically aligned representations of meteorological data and language, we introduce MeteoLIP, a dual-encoder framework that maps both modalities into a shared representation space. As meteorological inputs comprise 63 surface and atmospheric variables, MeteoLIP uses MetTok to aggregate variable-specific patch embeddings into compact meteorology tokens conditioned on temporal and geospatial information. During alignment training, MeteoLIP samples one of the four caption styles for each meteorological input, exposing the model to complementary descriptions of the same Earth system condition.

Our contributions can be summarized as follows: \textbf{(1)} We introduce \name{}, a large-scale Earth system--language dataset with broad geographic coverage, diverse natural phenomena and normal-weather conditions, metadata-grounded multi-style captions, and alignment with multivariate ERA5 data; \textbf{(2)} We introduce MeteoLIP for meteorology-language representation learning, and train Meteo-CLIP and Meteo-SigLIP variants that improve over their corresponding CLIP and SigLIP baselines; \textbf{(3)} We establish three evaluation tasks spanning cross-modal alignment, semantic understanding, and quantitative inference through retrieval, weather-condition classification, and physical variable estimation.}

\section{Related Work}

\textbf{Vision-language alignment.}
Vision-language pretraining learns shared representations by aligning visual data with rich natural language descriptions. CLIP~\citep{clip} demonstrated the effectiveness of large-scale image-text contrastive learning, while SigLIP~\citep{siglip} and InternVL~\citep{internvl} further improved its training objective and model scaling. Beyond natural images, this paradigm has been adapted to specialized domains such as medical imaging~\citep{biomedclip,unimedclip} and remote sensing~\citep{remoteclip,dofaclip}. Due to their transferable representations, these visual encoders are widely adopted in multimodal language models across domains~\citep{llava,internvl35,medgemma,earthdial}. Despite these developments, vision-language alignment for Earth system data remains underexplored, primarily due to the lack of large-scale language supervision paired with geolocation and time information. \name{} addresses this gap by introducing geocoded natural language captions describing diverse historical weather conditions and natural phenomena from around the world, enabling their alignment with Earth system data, such as gridded meteorological data.

\textbf{Multimodal weather and climate datasets and benchmarks.}
As summarized in Table~\ref{tab:related_work_comparison}, recent work has increasingly connected meteorological data with natural language for weather- and climate-related tasks. WeatherQA~\citep{weatherqa} pairs expert forecast analyses with mesoscale analysis images for severe-weather question answering. ClimateIQA~\citep{climateiqa} generates meteorological heatmaps from four ERA5 variables~\citep{era5paper} to detect three types of severe weather events. Aquilon~\citep{aquilon} defines seven question types and constructs samples from ERA5 and an expert-curated event record. CLLMate~\citep{cllmate} curates expert-verified weather and climate events from environmental news and pairs them with four ERA5 variables for event prediction. MeteorPred~\citep{meteorpred} uses meteorological agency warnings as supervision and ERA5 variables as inputs for warning generation in China. While these datasets vary in scope and task formulation, they are primarily designed for instruction following or question answering. In contrast, \name{} provides large-scale {dataset spanning 2000--2024, with broad geographic coverage  and natural phenomena}, with four grounded caption styles per record to enable vision-language alignment with Earth system data. We further introduce retrieval, classification, and physical variable estimation tasks to evaluate multimodal representations in this domain.

\section{Dataset and Evaluation Tasks}

\subsection{Dataset Construction}
\label{sec:data_construction}

\begin{figure}[t]
    \begin{center}
    \includegraphics[width=0.9\textwidth]{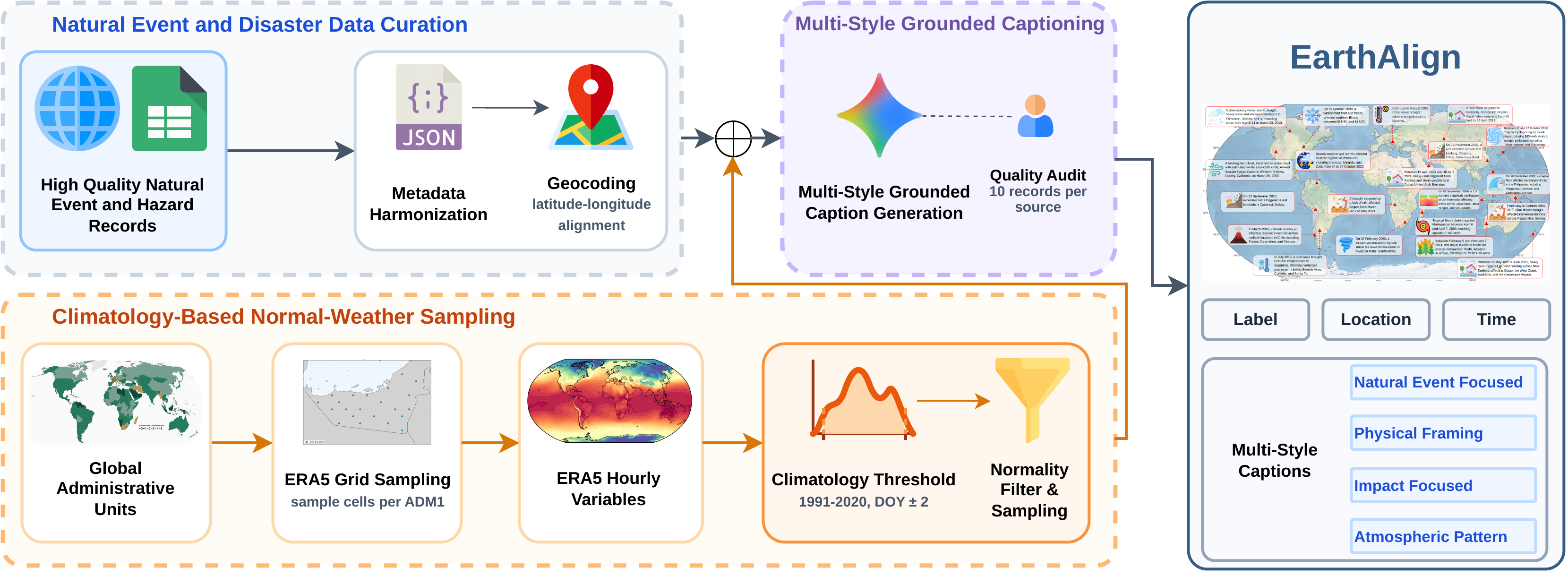}
    \end{center}
    \caption{\textbf{Data curation pipeline.}
    \name{} combines expert-curated natural-event records with climatology-based normal-weather samples. The resulting records contain standardized spatial and temporal information, labels, and four grounded caption styles for alignment with Earth system data.}
    \label{fig:data_curation_pipeline}
\end{figure}

To address the limitations of existing multimodal climate datasets, we present \name{}, a large-scale dataset designed to facilitate alignment of Earth-system data with language. We build the dataset based on two types of sources: records of natural events and disasters, and climatological data for normal weather samples. \name{} comprises \numall{} samples with {broad} geographic coverage spanning 2000--2024 and {encompasses a diverse set of keywords describing a wide range of natural phenomena} (Fig.~\ref{fig:data_distribution_and_statistics}). Each sample contains temporal information, a latitude-longitude bounding box, and four captions in different styles. Our data curation pipeline is shown in Fig.~\ref{fig:data_curation_pipeline}.

\textbf{Natural event and disaster records.} Natural events and disasters constitute the majority of samples in \name{}, providing large-scale and diverse observations of natural phenomena. These records were collected from 10 expert-curated sources (Appendix Table~\ref{tab:expert_event_sources}): 1) EM-DAT \citep{EM-DAT}, 2) NOAA Storm Event Database \citep{noaa_storm_events}, 3) IBTrACS \citep{ibtracs_paper,ibtracs_data}, 4) COOLR \citep{coolr}, 5) PIKART \citep{pikart}, 6) AusDE \citep{ausde}, 7) MexDE \citep{mexde}, 8) CanDE \citep{cande}, 9) IndDE \citep{indde}, and 10) FraDE \citep{frade}. Since these sources provide different types of information in different formats, we apply source-specific processing steps (Appendix~\ref{appendix:expert_curated_records}). In general, we begin by manually filtering the records to retain only natural phenomena. We then extract all available metadata to support the grounded generation of detailed descriptions. For sources that provide only textual location names, we apply geocoding to obtain coordinates and latitude-longitude bounding boxes. After processing, we obtain \numexpertrecord{} records with standardized phenomenon categories, associated metadata, as well as temporal and geospatial information.

\textbf{Climatology-based normal-weather sampling.}
To complement the natural event and disaster records, we identify normal-weather conditions relative to local climatology. We construct local climatologies from ERA5 \citep{era5paper} hourly variables over the 1991--2020 WMO reference period \citep{wmo_clim_30}. Adapting the centered 5-day window proposed in \cite{zhang2005avoiding}, we compute each variable-specific threshold for each day of year and hour using a centered $\pm 2$-day window. For scalability, we sample 20\% of the ERA5 grid cells within each first-level administrative unit (ADM1), based on administrative boundaries from geoBoundaries~\citep{geoboundaries}, with at least 10 cells whenever possible, and retrieve hourly variables from 1991 to 2025. Depending on the variable, we define normality by P05--P95, $\leq$P90, or $\geq$P10 thresholds. A grid-cell hour is considered normal only if all variables satisfy their respective thresholds, and an ADM1-hour is retained when at least 80\% of its sampled cells are normal. Using this conservative filter, we retain 1.6\% and 11\% of candidate hours in California and Abu Dhabi, respectively. We then sample eligible normal-weather hours globally over 2000--2024 with broad country coverage and area-proportional sampling, yielding \numnegsampling{} samples. Further details on the meteorological variables and their variable-specific normality criteria are provided in Appendix~\ref{app:normal_weather}.

\textbf{Grounded caption generation.}
From the metadata, we generate four stylistically different captions (i.e. \textit{natural-event-focused}, \textit{physical-framing}, \textit{impact-focused}, and \textit{atmospheric-pattern}) to provide complementary semantic perspectives of each record and support alignment with Earth system data. We use Gemini 3.1 Pro with a prompt that requires every caption to be directly supported by the input metadata and excludes unsupported causes, impacts, forecasts, and descriptive modifiers. To assess caption grounding and adherence to the target styles, a trained graduate annotator audited a stratified sample of 440 captions across all data sources and caption styles. The overall pass rate was 95.68\% (421/440), with style-specific pass rates of 100\%, 98\%, 89\%, and 95\%. We provide the generation prompt in Appendix~\ref{app:caption_generation}, and the audit rubric and error analysis in Appendix~\ref{app:caption_audit}.

\textbf{Earth system data alignment.}
Earth system data are heterogeneous across data types and modalities, variables, and spatial and time resolutions. ERA5 \citep{era5paper} provides globally gridded hourly data at $0.25^\circ$ resolution and is widely used in recent Earth system models and multimodal meteorological benchmarks \citep{weatherbench2,aurorafm,cllmate,meteorpred}. Motivated by this, we adopt ERA5 as a common baseline while preserving compatibility with other Earth system products. For each record, we extract a single multivariate ERA5 patch at the corresponding timestamp and spatial extent. Each patch contains 28 single-level variables and 7 pressure-level variables at 5 pressure levels, resulting in 63 input variables as listed in Appendix~\ref{app:era5_variables}.

\subsection{Evaluation Tasks}
To comprehensively evaluate Earth system multimodal representations, we devise three downstream tasks assessing their cross-modal alignment, semantic understanding, and quantitative inference capability: 
\textbf{(1) Retrieval} evaluates the alignment between Earth system (e.g. ERA5) representations and their corresponding textual descriptions using bidirectional text-to-meteorology (T2M) and meteorology-to-text (M2T) retrieval;
\textbf{(2) Classification} evaluates whether Earth system representations can capture category-level semantics of meteorological conditions. Given an Earth system input (e.g. ERA5), the model predicts one of 11 weather conditions: normal weather, drought, heat wave, cold wave, frost, atmospheric river, cyclone, convective storm, high wind, fog, and winter weather; and
\textbf{(3) Physical variable estimation} evaluates quantitative inference by estimating cyclone maximum sustained wind speed and minimum central pressure from an Earth system input.

\section{MeteoLIP: Aligning Meteorological Data with Language}

\begin{figure}[t]
    \begin{center}
    \includegraphics[width=0.9\textwidth]{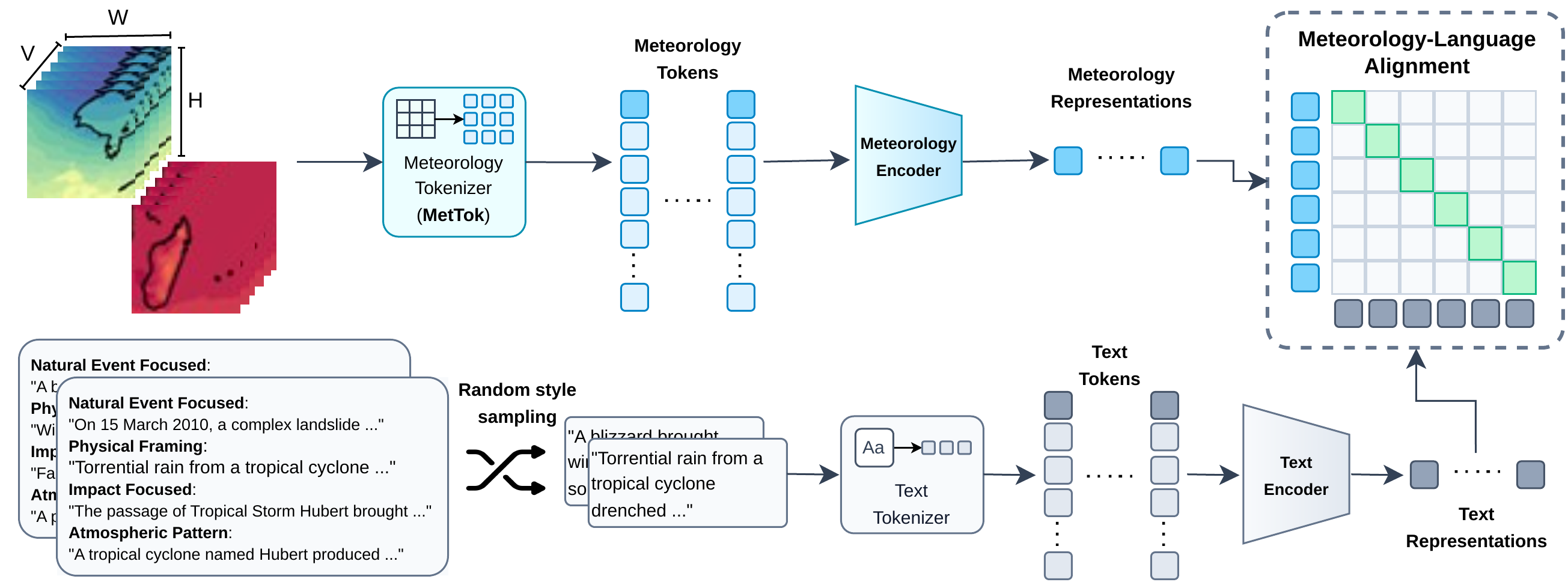}
    \end{center}
    \caption{\textbf{MeteoLIP Overview.}
    MeteoLIP consists of transformer-based meteorological and text encoders. MetTok converts multivariate meteorological input into a compact token sequence. During training, we randomly sample a caption style for each meteorological input and optimize a contrastive objective to align meteorological and textual representations in a shared embedding space.}
    \label{fig:meteolip}
\end{figure}

\name{} comprises paired ERA5 data and complementary textual descriptions. To learn aligned representations of these descriptions and their meteorological conditions, we introduce MeteoLIP, a dual-encoder framework for aligning meteorological data with natural language. As illustrated in Fig.~\ref{fig:meteolip}, MeteoLIP utilizes separate transformer-based meteorology and text encoders and maps their outputs into a shared representation space. To process the 63-variable ERA5 input, MeteoLIP uses MetTok, which provides a mechanism for incorporating geospatial and temporal information while producing a compact token sequence whose length depends on the spatial resolution. Through this alignment, MeteoLIP learns cross-modal representations that enable a range of downstream tasks.

\subsection{Model Architecture and Training Objective}

\textbf{Dual-encoder formulation.}
Let $\mathbf{x}_{T}$ denote a textual input and $\mathbf{x}_{M}\in\mathbb{R}^{V\times H\times W}$ denote meteorological data, where $V$ is the number of variables and $H\times W$ are the spatial dimensions. Each meteorological input is supplemented with temporal information $\mathbf{t}\in\mathbb{R}^{3}$, corresponding to year, day of year, and hour of day, and latitude--longitude geolocation information $\boldsymbol{\ell}\in\mathbb{R}^{2\times H\times W}$.
MeteoLIP employs dual encoders to map the two modalities into a shared $D$-dimensional representation space, with
$\mathbf{z}_{M}=\mathcal{F}_{M}\!\left(\mathcal{T}_{M}\!\left(\mathbf{x}_{M},\mathbf{t},\boldsymbol{\ell}\right)\right)$
and
$\mathbf{z}_{T}=\mathcal{F}_{T}\!\left(\mathcal{T}_{T}\!\left(\mathbf{x}_{T}\right)\right)$,
where $\mathbf{z}_{M},\mathbf{z}_{T}\in\mathbb{R}^{D}$ denote the meteorological and textual representations, respectively.
$\mathcal{T}_{M}$ and $\mathcal{T}_{T}$ denote the meteorology and text tokenizers, while $\mathcal{F}_{M}$ and $\mathcal{F}_{T}$ denote the transformer-based meteorology and text encoders.

\textbf{Meteorology tokenization (MetTok).}
Unlike natural images, meteorological inputs contain many physical variables. Let $ps$ be the patch size and $P=HW/{ps}^2$ be the number of patches. Independently patchifying each variable would produce $VP$ tokens, which would increase the self-attention complexity in the meteorology encoder from $\mathcal{O}(P^2D)$ to $\mathcal{O}(V^2P^2D)$. MetTok instead aggregates the variables within each patch, thereby retaining a sequence of $P$ tokens.
Specifically, we independently patchify each variable in $\mathbf{x}_M$ using a variable-specific linear projection and subsequently add a learnable variable-identity embedding, producing variable-patch embeddings $\mathbf{E}_M \in \mathbb{R}^{V \times P \times d}$, where $d=D/8$ is the MetTok embedding dimension. We encode the temporal information $\mathbf{t}$ by linearly normalizing the year and applying sine-cosine transformations to the day of year and hour of day, resulting in $\widetilde{\mathbf{t}} \in \mathbb{R}^{5}$, which is spatially expanded to $\mathbb{R}^{5 \times H \times W}$. Similarly, we apply sine-cosine transformations to latitude and longitude, resulting in $\widetilde{\boldsymbol{\ell}} \in \mathbb{R}^{4 \times H \times W}$.
To obtain temporal-geospatial patch embeddings $\mathbf{E}_{t\ell}\in\mathbb{R}^{P\times d}$, we concatenate the spatially expanded $\widetilde{\mathbf{t}}$ and $\widetilde{\boldsymbol{\ell}}$ along the channel dimension and jointly patchify them using a linear--GELU--linear projection, allowing temporal and geospatial features to interact nonlinearly.
Inspired by \cite{aurorafm,perceiver}, we introduce a learnable query embedding $\mathbf{e}_q \in \mathbb{R}^{d}$ to aggregate the variable embeddings in $\mathbf{E}_M$ through cross-attention. Since meteorological conditions depend on geographic location and temporal context, we condition the query on $\mathbf{E}_{t\ell}$ to obtain time- and location-aware queries $\mathbf{E}_q = [\mathbf{e}_q + \alpha \, \mathbf{e}_{t\ell,1}; \ldots; \mathbf{e}_q + \alpha \, \mathbf{e}_{t\ell,P}] \in \mathbb{R}^{P \times d}$, where $\alpha \in \mathbb{R}$ is a learnable scalar.
We then apply cross-attention independently within each patch, using the time- and location-aware query as the query and the corresponding variable embeddings as keys and values, yielding
$\mathbf{e}_{o,p} = \mathbf{e}_{q,p} + \operatorname{CrossAttn}(\mathbf{e}_{q,p}, \mathbf{e}_{M,p}, \mathbf{e}_{M,p})$
for $p=1,\ldots,P$, where $\mathbf{e}_{q,p}\in\mathbb{R}^{d}$ and $\mathbf{e}_{M,p}\in\mathbb{R}^{V\times d}$ are the query and variable embeddings for patch $p$, respectively.
We then project each aggregated representation to the meteorology encoder dimension using a linear layer, $\widehat{\mathbf{e}}_{o,p}=\mathbf{W}_o\mathbf{e}_{o,p}+\mathbf{b}_o\in\mathbb{R}^{D}$. The resulting meteorology tokens $\left[\widehat{\mathbf{e}}_{o,1}; \ldots; \widehat{\mathbf{e}}_{o,P}\right]$ are prepended with a class token and passed to the meteorology encoder, thereby avoiding the additional quadratic self-attention cost induced by variable-wise tokenization.

\textbf{Learning objective.}
During training, we sample one caption style for each meteorological input in a minibatch to expose the model to diverse linguistic descriptions from different perspectives and improve robustness. Given a minibatch of $B$ paired meteorological and textual inputs, we compute their pairwise cosine similarities as
$s_{ij}
=
\operatorname{sim}\!\left(
    \mathbf{z}_{M}^{(i)},
    \mathbf{z}_{T}^{(j)}
\right)
$.
Following \cite{clip,siglip}, we optimize a cross-modal contrastive objective that increases $s_{ii}$ for paired meteorological and textual representations while decreasing $s_{ij}$ for unmatched pairs $i\neq j$.

\subsection{Task-Specific Inference} \label{sec:task_specific_inference}

\textbf{Cross-modal retrieval.}
Given a retrieval set of $N$ paired samples
$\{(\mathbf{x}_{M}^{(i)},\mathbf{x}_{T}^{(i)})\}_{i=1}^{N}$,
MeteoLIP encodes each meteorological and textual input into shared representations
$\mathbf{z}_{M}^{(i)}$ and $\mathbf{z}_{T}^{(i)}$, respectively. We then compute the cosine similarity between the two modalities as
$s_{ij}=\operatorname{sim}\left(\mathbf{z}_{M}^{(i)},\mathbf{z}_{T}^{(j)}\right)$. For T2M retrieval, we rank meteorological candidates by their similarity to a textual query, and reverse the direction for M2T retrieval.

\textbf{Classification.}
We construct $K$ textual prompts $\{p_k(c)\}_{k=1}^{K}$ for each class $c\in\mathcal{C}$ and apply prompt ensembling to obtain a class prototype $\overline{\mathbf{z}}_{T}^{(c)}=\frac{1}{K}\sum_{k=1}^{K}\mathcal{F}_{T}\!\left(\mathcal{T}_{T}\!\left(p_k(c)\right)\right)$. Given a meteorological representation $\mathbf{z}_{M}$, we predict its class as $\hat{c}=\arg\max_{c\in\mathcal{C}}\operatorname{sim}\!\left(\mathbf{z}_{M},\overline{\mathbf{z}}_{T}^{(c)}\right)$.

\textbf{Physical variable estimation.}
Inspired by \cite{echoclip,fetalclip}, we estimate cyclone maximum sustained wind speed and minimum central pressure from $\mathbf{x}_{M}$ using value-conditioned text prototypes. We begin by constructing $K=80$ textual templates $\{p_k(v)\}_{k=1}^{K}$, such as \textit{``a cyclone with sustained wind speeds of $v$ knots''} for each candidate value $v\in\mathcal{V}$, where $\mathcal{V}=\{v_{\min},v_{\min}+1,\ldots,v_{\max}\}$, and obtain a value prototype $\overline{\mathbf{z}}_{T}^{(v)}=\frac{1}{K}\sum_{k=1}^{K}\mathcal{F}_{T}\!\left(\mathcal{T}_{T}\!\left(p_k(v)\right)\right)$. We then compute the similarity score for each $v$ as $s_v(\mathbf{z}_M)=\operatorname{sim}\left(\mathbf{z}_{M},\overline{\mathbf{z}}_{T}^{(v)}\right)$. Let $\mathcal{S}_\rho(\mathbf{x}_M)$ denote the top-$\rho$ candidate values by score, where $\rho$ is the nearest odd integer to $0.05|\mathcal{V}|$. For robust estimation, we take the median of $\mathcal{S}_\rho(\mathbf{x}_M)$ as the final estimate.

\section{Experiments and Results}

\subsection{Experimental Setup} \label{sec:experimental_setup}

\textbf{Meteorological data processing.} We use the 63 ERA5 meteorological variables listed in Appendix~\ref{app:era5_variables}. We use the specified time for records with an exact timestamp, while we sample a timestamp within the reported interval for records spanning a time range. We standardize each meteorological input to $56\times56$ grid cells. For samples spanning less than $14^\circ\times14^\circ$, we expand the bounding box to this extent. For larger samples, we construct a square region using the larger latitudinal or longitudinal extent and linearly interpolate it to $56\times56$ to avoid introducing extremes. We also normalize each variable using its mean and standard deviation computed across samples.

\textbf{Dataset split.} We employ a year-based train-test split to assess whether the learned representations remain effective across years, using 2000--2010 and 2014--2024 for training and withholding 2011--2013 for evaluation on unseen intermediate years. This results in \numtrain{} training samples, with \numtestretrieval{}, \numtestclassification{}, \numtestwind{}, and \numtestpres{} test samples for retrieval, classification, cyclone maximum sustained wind speed estimation, and cyclone minimum central pressure estimation, respectively.

\textbf{MeteoLIP and training configuration.} We investigate two MeteoLIP variants based on the CLIP~\citep{clip} and SigLIP~\citep{siglip} architectures and their respective learning objectives, denoted as Meteo-CLIP and Meteo-SigLIP. We initialize the encoders from the corresponding OpenAI CLIP and SigLIP WebLI checkpoints and use a text context length of 77 tokens.
For the meteorology encoder, we use a patch size of $4\times4$, which results in $(56/4)^2=196$ spatial patches.
Unless otherwise specified, we train all models for 30 epochs on 4$\times$A100 (40GB) GPUs with a batch size of 256 per GPU, using the AdamW optimizer with a learning rate of $1 \times 10^{-4}$ and weight decay of $1 \times 10^{-2}$. We also apply a cosine learning-rate schedule with 2,000 warmup steps.
We report the mean and standard deviation over three training runs using the final epoch checkpoint.

\textbf{Baselines.} For the CLIP and SigLIP baselines, following the multichannel patchification used in FourCastNet~\citep{fourcastnet}, we concatenate all meteorological variables along the channel dimension and jointly project each spatial patch using a linear layer, producing one token per patch.
To follow the original image-text pair formulation of CLIP and SigLIP, we train the baselines using only the \textit{natural-event-focused} caption, which identifies the natural phenomenon together with its temporal and geospatial context when available.

\subsection{Main Results}

\textbf{Retrieval.} To evaluate meteorology-language alignment, we first assess bidirectional cross-modal retrieval. As shown in Table~\ref{tab:retrieval_results}, Meteo-CLIP and Meteo-SigLIP improve over their corresponding CLIP and SigLIP baselines across all retrieval metrics. Meteo-CLIP improves M2T and T2M R@1 from 20.25 and 20.03 to 58.79 and 60.10 respectively, while Meteo-SigLIP achieves the best performance with 66.07 and 66.79. Similar improvements are observed at R@5 and R@10, demonstrating strong meteorology-language alignment. Since the default baselines use a single caption style, we further examine the effects of multi-style supervision and MetTok conditioning in Section~\ref{sec:ablation}.

\textbf{Classification.}
Table~\ref{tab:classification_results} evaluates category-level meteorological representations under two protocols: \textit{text-prototype inference}, which applies the text-based inference described in Section~\ref{sec:task_specific_inference} without task-specific post-training, and \textit{linear probing}, which trains a linear prediction head on top of the frozen meteorology encoder. Under text-prototype inference, Meteo-CLIP improves F1 from 30.15 to 47.91 and Top-1 accuracy from 33.69 to 57.69, while Meteo-SigLIP improves from 21.56 to 50.19 and Top-1 accuracy from 22.13 to 55.59. The improvements can also be observed under linear probing, where Meteo-SigLIP achieves the highest F1 of 80.75 and Top-1 accuracy of 92.35. These results indicate that MeteoLIP improves both language-based inference and the category-level semantic information encoded in the frozen meteorology representations. We provide detailed category-wise results in Appendix~\ref{app:classification_category_result}, where drought, heat wave, and cold wave are identified as the most challenging categories.

\textbf{Physical variable estimation.}
As provided in Table~\ref{tab:physical_results}, under text-prototype inference, Meteo-CLIP improves $R^2$ for cyclone maximum sustained wind speed from 5.29 to 62.27 and for minimum central pressure from 26.81 to 67.64, while Meteo-SigLIP improves them from $-11.59$ to 60.17 and from 45.12 to 73.39, respectively. Under linear probing, Meteo-SigLIP achieves the best performance for both variables, with $R^2$ values of 71.85 for maximum sustained wind speed and 77.36 for minimum central pressure.
We further compare against physical baselines derived directly from ERA5 in Appendix~\ref{app:physical_baselines}.
These results demonstrate that the learned meteorology-language representations are capable of capturing physically meaningful quantitative information.

\begin{table}[t]
\caption{\textbf{Comparison of CLIP and SigLIP baselines with their MeteoLIP counterparts.}
(a) Cross-modal retrieval.
(b) Classification under \textit{text-prototype inference} and \textit{linear probing}.
(c) Physical variable estimation of cyclone maximum sustained wind speed and minimum central pressure under the same two protocols.
\textbf{Bold} and \underline{underline} indicate the best and second-best results.}
\label{tab:main_results}
\centering
\small
\setlength{\tabcolsep}{4pt}
\renewcommand{\arraystretch}{1.05}

\begin{subtable}[t]{\linewidth}
\phantomsubcaption
\label{tab:retrieval_results}
\centering

\begin{adjustbox}{max width=0.8\linewidth}
\begin{tabular}{@{}lcccccc@{}}
\toprule
\multicolumn{7}{@{}l}{\textbf{(a)}} \\
& \multicolumn{3}{c}{\textbf{Meteorology-to-Text}}
& \multicolumn{3}{c}{\textbf{Text-to-Meteorology}} \\
\cmidrule(lr){2-4} \cmidrule(lr){5-7}

\textbf{Model}
& \textbf{R@1} $\uparrow$
& \textbf{R@5} $\uparrow$
& \textbf{R@10} $\uparrow$
& \textbf{R@1} $\uparrow$
& \textbf{R@5} $\uparrow$
& \textbf{R@10} $\uparrow$ \\
\midrule

CLIP
& $20.25 \pm 0.13$
& $45.67 \pm 0.19$
& $58.25 \pm 0.27$
& $20.03 \pm 0.14$
& $45.05 \pm 0.39$
& $57.43 \pm 0.47$ \\

\rowcolor{TPIMacro} Meteo-CLIP
& \underline{$58.79 \pm 1.43$}
& \underline{$88.13 \pm 0.60$}
& \underline{$94.09 \pm 0.32$}
& \underline{$60.10 \pm 0.23$}
& \underline{$87.68 \pm 0.13$}
& \underline{$93.50 \pm 0.12$} \\

SigLIP
& $32.06 \pm 0.56$
& $61.11 \pm 0.75$
& $72.07 \pm 0.69$
& $30.23 \pm 0.22$
& $58.12 \pm 0.25$
& $68.98 \pm 0.25$ \\

\rowcolor{TPIMacro} Meteo-SigLIP
& $\mathbf{66.07 \pm 0.64}$
& $\mathbf{91.77 \pm 0.29}$
& $\mathbf{96.14 \pm 0.15}$
& $\mathbf{66.79 \pm 0.50}$
& $\mathbf{91.27 \pm 0.31}$
& $\mathbf{95.62 \pm 0.18}$ \\

\bottomrule
\end{tabular}
\end{adjustbox}
\end{subtable}

\begin{subtable}[t]{\linewidth}
\phantomsubcaption
\label{tab:classification_results}
\centering

\begin{adjustbox}{max width=0.8\linewidth}
\begin{tabular}{@{}lcccccc@{}}
\toprule
\multicolumn{7}{@{}l}{\textbf{(b)}} \\
& \multicolumn{3}{c}{\textbf{Text-Prototype Inference}}
& \multicolumn{3}{c}{\textbf{Linear Probing}} \\
\cmidrule(lr){2-4} \cmidrule(lr){5-7}

\textbf{Model}
& \textbf{F1} $\uparrow$
& \textbf{Top-1 Acc.} $\uparrow$
& \textbf{Top-3 Acc.} $\uparrow$
& \textbf{F1} $\uparrow$
& \textbf{Top-1 Acc.} $\uparrow$
& \textbf{Top-3 Acc.} $\uparrow$ \\
\midrule

CLIP
& $30.15 \pm 2.83$
& $33.69 \pm 3.10$
& $66.95 \pm 2.53$
& $75.04 \pm 0.20$
& $90.61 \pm 0.05$
& $98.62 \pm 0.06$ \\

\rowcolor{TPIMacro} Meteo-CLIP
& \underline{$47.91 \pm 2.10$}
& $\mathbf{57.69 \pm 1.43}$
& $\mathbf{85.98 \pm 0.97}$
& \underline{$79.98 \pm 0.26$}
& \underline{$92.16 \pm 0.11$}
& $\mathbf{98.82 \pm 0.05}$ \\

SigLIP
& $21.56 \pm 1.48$
& $22.13 \pm 2.95$
& $50.15 \pm 3.52$
& $76.30 \pm 0.54$
& $91.06 \pm 0.08$
& $98.61 \pm 0.05$ \\

\rowcolor{TPIMacro} Meteo-SigLIP
& $\mathbf{50.19 \pm 2.95}$
& \underline{$55.59 \pm 5.31$}
& \underline{$82.29 \pm 3.63$}
& $\mathbf{80.75 \pm 0.14}$
& $\mathbf{92.35 \pm 0.04}$
& \underline{$98.78 \pm 0.04$} \\

\bottomrule
\end{tabular}
\end{adjustbox}
\end{subtable}

\begin{subtable}[t]{\linewidth}
\phantomsubcaption
\label{tab:physical_results}
\centering

\begin{adjustbox}{max width=0.8\linewidth}
\begin{tabular}{@{}llcccccc@{}}
\toprule
\multicolumn{8}{@{}l}{\textbf{(c)}} \\
& &
\multicolumn{3}{c}{\textbf{Text-Prototype Inference}} &
\multicolumn{3}{c}{\textbf{Linear Probing}} \\
\cmidrule(lr){3-5} \cmidrule(lr){6-8}

\textbf{Variable} &
\textbf{Model} &
$R^2 \uparrow$ &
\textbf{MAE}$\downarrow$ &
\textbf{RMSE}$\downarrow$ &
$R^2 \uparrow$ &
\textbf{MAE}$\downarrow$ &
\textbf{RMSE}$\downarrow$ \\
\midrule

\multirow{4}{*}{\shortstack[l]{\textbf{Sustained}\\\textbf{Wind}}}
& CLIP
& $5.29 \pm 19.85$
& $14.62 \pm 1.38$
& $21.38 \pm 2.24$
& $68.09 \pm 1.04$
& $9.42 \pm 0.17$
& $12.46 \pm 0.20$ \\

\rowcolor{TPIMacro} & Meteo-CLIP
& $\mathbf{62.27 \pm 4.75}$
& \underline{$9.35 \pm 0.42$}
& $\mathbf{13.53 \pm 0.84}$
& \underline{$71.34 \pm 0.51$}
& \underline{$8.65 \pm 0.08$}
& \underline{$11.80 \pm 0.10$} \\

& SigLIP
& $-11.59 \pm 16.49$
& $14.46 \pm 0.88$
& $23.25 \pm 1.72$
& $70.52 \pm 0.48$
& $8.91 \pm 0.11$
& $11.97 \pm 0.10$ \\

\rowcolor{TPIMacro} & Meteo-SigLIP
& \underline{$60.17 \pm 1.73$}
& $\mathbf{9.32 \pm 0.21}$
& \underline{$13.91 \pm 0.30$}
& $\mathbf{71.85 \pm 0.37}$
& $\mathbf{8.50 \pm 0.04}$
& $\mathbf{11.70 \pm 0.08}$ \\

\midrule

\multirow{4}{*}{\shortstack[l]{\textbf{Central}\\\textbf{Pressure}}}
& CLIP
& $26.81 \pm 14.46$
& $9.70 \pm 0.80$
& $16.54 \pm 1.60$
& $74.08 \pm 0.82$
& $7.04 \pm 0.11$
& $9.87 \pm 0.16$ \\

\rowcolor{TPIMacro} & Meteo-CLIP
& \underline{$67.64 \pm 4.23$}
& \underline{$6.37 \pm 0.22$}
& \underline{$11.01 \pm 0.72$}
& \underline{$76.36 \pm 0.46$}
& \underline{$6.41 \pm 0.09$}
& \underline{$9.43 \pm 0.09$} \\

& SigLIP
& $45.12 \pm 3.00$
& $7.96 \pm 0.20$
& $14.36 \pm 0.39$
& $75.98 \pm 0.35$
& $6.60 \pm 0.06$
& $9.50 \pm 0.07$ \\

\rowcolor{TPIMacro} & Meteo-SigLIP
& $\mathbf{73.39 \pm 1.61}$
& $\mathbf{5.96 \pm 0.09}$
& $\mathbf{10.00 \pm 0.30}$
& $\mathbf{77.36 \pm 0.45}$
& $\mathbf{6.16 \pm 0.09}$
& $\mathbf{9.23 \pm 0.09}$ \\

\bottomrule
\end{tabular}
\end{adjustbox}
\end{subtable}

\end{table}

\subsection{Ablation Studies} \label{sec:ablation}

\textbf{Caption supervision.}
We study the effects of caption style for retrieval and multi-style supervision. Table~\ref{tab:caption_style_ablation} shows that the natural-event-focused caption results in the highest retrieval performance, achieving 58.79 M2T and 60.12 T2M R@1. In comparison, the other caption styles yield substantially lower retrieval performance, suggesting that explicitly describing the natural phenomenon together with its temporal and geospatial context provides more instance-discriminative information for retrieval. Table~\ref{tab:caption_number_ablation} further demonstrates the benefit of multi-style supervision during training. Increasing from one to four caption styles improves both model architectures: CLIP increases from 20.25/20.03 to 31.64/31.10 M2T/T2M R@1, while Meteo-CLIP improves from 34.24/55.55 to 58.79/60.10. Moreover, Meteo-CLIP consistently outperforms CLIP under both one-style and four-style training, demonstrating its effectiveness across different levels of caption-style supervision.

\textbf{Model initialization.}
As shown in Table~\ref{tab:initialization_ablation}, initializing from the OpenAI CLIP checkpoint improves M2T R@1 from 34.46 to 58.79 and T2M R@1 from 41.43 to 60.10 compared with training from scratch. This suggests that, despite the modality gap between meteorological data and RGB images, pretrained visual representations provide a useful initialization for learning spatial structure in meteorological fields.
One possible explanation is that some meteorological variables can reflect coherent spatial patterns associated with natural phenomena such as cyclones and atmospheric rivers, which can benefit from the spatial pattern recognition capabilities of pretrained image encoders.

{
\textbf{MetTok conditioning.}
Table~\ref{tab:mettok_components} highlights the contribution of meteorological input and MetTok's conditioning components across retrieval, weather condition classification, and maximum sustained wind estimation. For retrieval, variable-identity embedding and geolocation provide relatively modest improvements, whereas temporal conditioning contributes substantially. Notably, using geolocation and time alone without meteorological input achieves 50.41/58.49 M2T/T2M R@1, showing that temporal and geographic metadata provide strong cues for matching natural event captions. However, this metadata-only configuration achieves substantially lower performance on weather condition classification and fails to provide useful sustained wind estimates, achieving 32.64 F1 and an $R^2$ of $-155.14$, compared with 47.91 F1 and 62.27 $R^2$ for the full MetTok configuration. To further examine the role of metadata in retrieval, Table~\ref{tab:retrieval_candidate_pool} compares the full MetTok configuration with metadata only using original and time-removed captions under different candidate pools. With the original captions, the relative gap between the two configurations is 14.3\% under both candidate pools. When temporal information is removed from the captions, the gap increases substantially: the full configuration achieves 15.41 M2T R@1 compared with 6.57 using metadata only over all test samples, and 38.65 compared with 23.26 when candidates are restricted to the same calendar month. Qualitative examples in Figure~\ref{fig:mettok_qualitative_retrieval} further show that the full configuration retrieves meteorologically related events, whereas using metadata alone can retrieve unrelated phenomena.
These results and analysis highlight the importance of meteorological input for semantic and physical inference and show that it also contributes substantially to retrieval when direct temporal cues are removed.
}

\begin{table}[t]
\centering

\begin{minipage}[t]{0.39\linewidth}
\centering

\caption{Meteo-CLIP retrieval performance by caption style.}
\label{tab:caption_style_ablation}

\small
\setlength{\tabcolsep}{3.5pt}
\renewcommand{\arraystretch}{1.08}

\begin{adjustbox}{max width=\linewidth}
\begin{tabular}{@{}lcc@{}}
\toprule
\textbf{Caption Style}
& \textbf{M2T R@1}$\uparrow$
& \textbf{T2M R@1}$\uparrow$ \\
\midrule

\rowcolor{TPIMacro}
Natural Event Focused
& $\mathbf{58.79 \pm 1.44}$
& $\mathbf{60.12 \pm 0.25}$ \\

Physical Framing
& $21.18 \pm 0.18$
& $21.67 \pm 0.08$ \\

Impact Focused
& $19.09 \pm 0.17$
& $18.55 \pm 0.08$ \\

Atmospheric Pattern
& $22.24 \pm 0.41$
& $21.52 \pm 0.08$ \\

\bottomrule
\end{tabular}
\end{adjustbox}

\end{minipage}
\hfill
\begin{minipage}[t]{0.28\linewidth}
\centering

\caption{Ablation on multi-style caption supervision during training. \textbf{\#}: \# Styles.}
\label{tab:caption_number_ablation}

\small
\setlength{\tabcolsep}{3.5pt}
\renewcommand{\arraystretch}{1.08}

\begin{adjustbox}{max width=\linewidth}
\begin{tabular}{@{}lccc@{}}
\toprule
\textbf{Model}
& \textbf{\#}
& \textbf{M2T R@1}$\uparrow$
& \textbf{T2M R@1}$\uparrow$ \\
\midrule

& 1
& $20.25 \pm 0.13$
& $20.03 \pm 0.14$ \\

\multirow{-2}{*}{CLIP}
& 4
& $\mathbf{31.64 \pm 0.24}$
& $\mathbf{31.10 \pm 0.05}$ \\

\midrule

& 1
& $34.24 \pm 8.52$
& $55.55 \pm 4.57$ \\

\rowcolor{TPIMacro}
\multirow{-2}{*}{Meteo-CLIP}
& 4
& $\mathbf{58.79 \pm 1.43}$
& $\mathbf{60.10 \pm 0.23}$ \\

\bottomrule
\end{tabular}
\end{adjustbox}

\end{minipage}
\hfill
\begin{minipage}[t]{0.30\linewidth}
\centering

\caption{Ablation on model initialization for Meteo-CLIP training.}
\label{tab:initialization_ablation}

\small
\setlength{\tabcolsep}{6pt}
\renewcommand{\arraystretch}{1.08}

\begin{adjustbox}{max width=\linewidth}
\begin{tabular}{@{}lcc@{}}
\toprule
\textbf{Init.}
& \textbf{M2T R@1}$\uparrow$
& \textbf{T2M R@1}$\uparrow$ \\
\midrule

Scratch
& $34.46 \pm 1.78$
& $41.43 \pm 2.21$ \\

\rowcolor{TPIMacro} OpenAI
& $\mathbf{58.79 \pm 1.43}$
& $\mathbf{60.10 \pm 0.23}$ \\

\bottomrule
\end{tabular}
\end{adjustbox}
\end{minipage}

\end{table}

\begin{table}[t]
\centering

\begin{minipage}[t]{0.65\linewidth}
\centering
\caption{
{Ablation of MetTok components.}
We train Meteo-CLIP under different MetTok configurations to assess the contribution of meteorological input, variable-identity embedding, geolocation, and temporal conditioning.
\textbf{Bold} and \underline{underlined} values indicate the best and second-best results, respectively.
}
\label{tab:mettok_components}
\small
\setlength{\tabcolsep}{4.0pt}
\renewcommand{\arraystretch}{1.10}
\begin{adjustbox}{max width=\textwidth}
\begin{tabular}{@{}cccccccccc@{}}
\toprule
\multicolumn{4}{c}{\textbf{Input \& Encoding}} & \multicolumn{2}{c}{\textbf{Retrieval}} & \multicolumn{2}{c}{\textbf{Weather Condition}} & \multicolumn{2}{c}{\textbf{Sustained Wind}} \\
\cmidrule(lr){1-4} \cmidrule(lr){5-6} \cmidrule(lr){7-8} \cmidrule(lr){9-10}
\textbf{Met.} & \textbf{Var. ID} & \textbf{Geo.} & \textbf{Time} & \textbf{M2T R@1} $\uparrow$ & \textbf{T2M R@1} $\uparrow$ & \textbf{F1} $\uparrow$ & \textbf{Acc.} $\uparrow$ & \textbf{$R^2$} $\uparrow$ & \textbf{MAE} $\downarrow$ \\
\midrule
$\checkmark$ & -- & -- & -- & $32.02 \pm 0.33$ & $31.74 \pm 0.19$ & $47.04 \pm 3.67$ & $56.01 \pm 6.49$ & $52.66 \pm 5.20$ & $10.33 \pm 0.40$ \\
$\checkmark$ & $\checkmark$ & -- & -- & $32.44 \pm 0.34$ & $32.71 \pm 0.21$ & $\mathbf{48.17 \pm 1.56}$ & \underline{$57.23 \pm 2.32$} & $46.74 \pm 13.37$ & $10.44 \pm 0.67$ \\
-- & -- & $\checkmark$ & $\checkmark$ & $50.41 \pm 2.43$ & $58.49 \pm 1.22$ & $32.64 \pm 4.30$ & $45.12 \pm 11.06$ & $-155.14 \pm 34.45$ & $25.75 \pm 1.93$ \\
$\checkmark$ & $\checkmark$ & $\checkmark$ & -- & $33.21 \pm 0.69$ & $32.83 \pm 0.44$ & $47.25 \pm 1.41$ & $54.93 \pm 3.20$ & $\mathbf{62.51 \pm 3.87}$ & \underline{$9.43 \pm 0.15$} \\
$\checkmark$ & $\checkmark$ & -- & $\checkmark$ & \underline{$57.30 \pm 1.65$} & \underline{$59.17 \pm 0.44$} & $41.27 \pm 1.19$ & $47.47 \pm 4.88$ & $59.82 \pm 4.56$ & $9.79 \pm 0.47$ \\
\rowcolor{TPIMacro} $\checkmark$ & $\checkmark$ & $\checkmark$ & $\checkmark$ & $\mathbf{58.79 \pm 1.43}$ & $\mathbf{60.10 \pm 0.23}$ & \underline{$47.91 \pm 2.10$} & $\mathbf{57.69 \pm 1.43}$ & \underline{$62.27 \pm 4.75$} & $\mathbf{9.35 \pm 0.42}$ \\
\bottomrule
\end{tabular}
\end{adjustbox}
\end{minipage}
\hfill
\begin{minipage}[t]{0.34\linewidth}

\centering
\caption{{Retrieval cue analysis.} M2T R@1 of Full MetTok and metadata only using original and time-removed captions under all test and same month candidate pools.}
\label{tab:retrieval_candidate_pool}

\definecolor{GapLow}{HTML}{EAF3F8}
\definecolor{GapMid}{HTML}{CFE4F1}
\definecolor{GapHigh}{HTML}{A9CEE5}

\small
\setlength{\tabcolsep}{5pt}
\renewcommand{\arraystretch}{1.08}

\begin{adjustbox}{max width=\linewidth}
\begin{tabular}{@{}llccc@{}}
\toprule
\textbf{Candidate Pool}
& \textbf{Caption}
& \textbf{Full MetTok}
& \textbf{Metadata only}
& \textbf{Rel. Gap} \\
\midrule

\multirow{2}{*}{All test samples}
& Original
& $58.79 \pm 1.43$
& $50.41 \pm 2.43$
& \cellcolor{GapLow}$14.3\%$ \\

& Time removed
& $15.41 \pm 0.02$
& $6.57 \pm 0.18$
& \cellcolor{GapHigh}$57.4\%$ \\

\midrule

\multirow{2}{*}{Same month}
& Original
& $61.84 \pm 2.48$
& $53.02 \pm 3.41$
& \cellcolor{GapLow}$14.3\%$ \\

& Time removed
& $38.65 \pm 4.06$
& $23.26 \pm 2.99$
& \cellcolor{GapMid}$39.8\%$ \\

\bottomrule
\end{tabular}
\end{adjustbox}
\end{minipage}

\end{table}

\textbf{Training data scaling.}
We study the effect of training data scale using 10\%, 25\%, 50\%, and 100\% of the training set. Figure~\ref{fig:training_data_scaling_abl} demonstrates that performance generally improves across all tasks with more training data, with the full training set achieving the best performance. The improvements are particularly notable for retrieval and maximum sustained wind speed estimation, while classification and minimum central pressure estimation also benefit from increased data scale. These results suggest that the model can further benefit from additional meteorology-language supervision.

\noindent
\begin{minipage}[t]{0.67\linewidth}
\vspace{0pt}
{
\textbf{Source-held-out generalization.}
We study source-level generalization by training Meteo-CLIP while simultaneously removing AusDE, CanDE, FraDE, IndDE, and MexDE from training, and then evaluating retrieval on the test sets of the five held-out record sources. Table~\ref{tab:ablation_5sources_holdout} shows that the decrease in retrieval performance varies substantially across sources, which may reflect differences in source-specific reporting criteria, local terminology, and spatial granularity. These results highlight the importance of local record sources for meteorology-language alignment, while showing that cross-source generalization remains challenging.
}
\end{minipage}
\hfill
\begin{minipage}[t]{0.31\linewidth}
\vspace{0pt}
\centering

\captionof{table}{Source-held-out M2T R1; mean over 3 runs.}
\label{tab:ablation_5sources_holdout}

\begin{adjustbox}{max width=0.8\linewidth}

\begin{tabular}{@{}lcc@{}}
\toprule
& \multicolumn{2}{c}{\textbf{Training Data}} \\
\cmidrule(lr){2-3}
\textbf{Test Data} & \textbf{Full} & \textbf{Held-Out} \\
\midrule
AusDE {\scriptsize\textit{(n=27)}} & $80.2$ & $72.8$ {\scriptsize\textcolor{red}{(-7.4)}} \\
CanDE {\scriptsize\textit{(n=47)}} & $64.5$ & $63.1$ {\scriptsize\textcolor{red}{(-1.4)}} \\
FraDE {\scriptsize\textit{(n=966)}} & $24.0$ & $10.6$ {\scriptsize\textcolor{red}{(-13.4)}} \\
IndDE {\scriptsize\textit{(n=3,835)}} & $69.6$ & $13.1$ {\scriptsize\textcolor{red}{(-56.5)}} \\
MexDE {\scriptsize\textit{(n=336)}} & $48.1$ & $37.5$ {\scriptsize\textcolor{red}{(-10.6)}} \\
\bottomrule
\end{tabular}
\end{adjustbox}

\end{minipage}

{
\section{Conclusion}
We present \name{}, a large-scale Earth system--language dataset that pairs meteorological data with multi-style natural language descriptions of diverse natural phenomena and normal-weather conditions. Alongside \name{}, we develop MeteoLIP to learn language-aligned representations of high-dimensional meteorological data while incorporating variable-specific, temporal, and geospatial information. Across cross-modal retrieval, weather-condition classification, and physical variable estimation, MeteoLIP improves over the corresponding CLIP and SigLIP baselines. Our analyses further demonstrate the importance of meteorological input, temporal and geospatial conditioning, multi-style language supervision, and diverse training data for effective meteorology-language alignment. We believe \name{} can provide a useful foundation for advancing language-aligned multimodal representation learning of Earth system data.
}

\subsection*{Acknowledgments}
This material is based on work supported by ADIA Lab, Abu Dhabi under a collaborative research project titled Climate Co-Pilot: A Multimodal AI Agent for Climate Resilience in the MENA Region.

{
\subsection*{AI use statement}

In this work, we used generative AI tools to generate captions from available metadata, assist with coding for experiments and visualizations, help identify record data sources and relevant literature, assist with translation when working with source-specific records in Mexico, France, and Indonesia, and provide writing assistance to improve clarity and readability. We have reviewed the AI-assisted work. For example, a subset of the generated captions was manually audited for consistency with the underlying metadata across the record data sources; AI-assisted code was manually inspected and executed; identified record data sources and literature were independently verified; and AI-assisted writing refinements were manually reviewed and further revised. We take responsibility for the final content of this work, including text, claims, or artifacts produced with the aid of generative AI.
}

{
\subsection*{Reproducibility statement}

The preprocessed data, source code, and model weights are publicly available at \href{https://github.com/mbzuai-oryx/EarthAlign}{https://github.com/mbzuai-oryx/EarthAlign}. The distribution of \name{} is reported in Figure~\ref{fig:data_distribution_and_statistics} and Appendix~\ref{app:data_statistics_and_distributions}, with representative examples illustrated in Figure~\ref{fig:representative_examples_global_map}. The overall dataset construction is described in Section~\ref{sec:data_construction}, while Appendices~\ref{app:dataset_samples_record_construction}, \ref{app:caption_generation_and_audit}, and~\ref{app:era5_variables} provide further details on source-specific record processing, grounded multi-style caption generation, and ERA5 meteorological data alignment and preprocessing, respectively. MeteoLIP model configurations and training hyperparameters are provided in Section~\ref{sec:experimental_setup}. We report experimental results as mean and standard deviation across independent runs to account for variation across training runs.
}

\bibliography{iclr2027_conference}
\bibliographystyle{iclr2027_conference}

\clearpage

\appendix

\section*{Appendix}
\addcontentsline{toc}{section}{Appendix}

\begin{table*}[h]
\centering
\caption{Comparison of recent existing multimodal weather and climate datasets by temporal and geographic coverage, evidence source, natural-phenomenon types, dataset size, and evaluation tasks.}

\scriptsize
\setlength{\tabcolsep}{3pt}
\renewcommand{\arraystretch}{1.05}

\begin{tabularx}{\textwidth}{
    @{}
    L{1.40cm}   
    C{1.15cm}   
    L{1.00cm}   
    L{2.05cm}   
    C{0.65cm}   
    C{1.65cm}   
    Y           
    @{}
}
\toprule
\textbf{Dataset} &
\textbf{Years} &
\textbf{Geo. Scope} &
\textbf{Evidence Source} &
\textbf{\# NP Types} &
\textbf{Size} &
\textbf{Tasks} \\
\midrule

WeatherQA
& 2014--2020
& CONUS
& Expert forecast analyses
& 7
& 8,511
& Severe weather MC-VQA and mesoscale case study generation \\

ClimateIQA
& 2023
& Global
& Meteorological heatmaps
& 3
& 26.3k img. (762.1k QA)
& Meteorological anomaly VQA and description generation \\

Aquilon
& 1979--2022
& Global
& Expert-curated records \& ERA5
& N/R
& 444,244
& Weather analysis, forecasting, and extreme event VQA \\

CLLMate
& 2015--2023
& Global
& Expert-verified news
& 13
& 7,747
& Weather and climate phenomena and consequences MC-VQA \\

MeteorPred
& 2023--2024
& China
& Agency warnings
& 7
& 421,363
& Severe weather VQA, MC-VQA, and structured generation tasks \\

\midrule
\rowcolor[RGB]{235,247,250}
\textbf{\name{}}
& 2000--2024
& Global
& Expert-curated records \& ERA5
& \numnpkeywords{}
& \numall{}
& VL alignment: retrieval, classification, and physical variable estimation \\

\bottomrule
\end{tabularx}
\label{tab:related_work_comparison}

\raggedright\footnotesize
\textbf{NP}: Natural phenomena.
\textbf{N/R}: not reported or unknown.
\textbf{VQA}: visual question answering.
\textbf{MC}: multiple-choice.
\textbf{VL}: vision-language.

\end{table*}

\begin{figure}[h]
    \begin{center}
    \includegraphics[width=\textwidth]{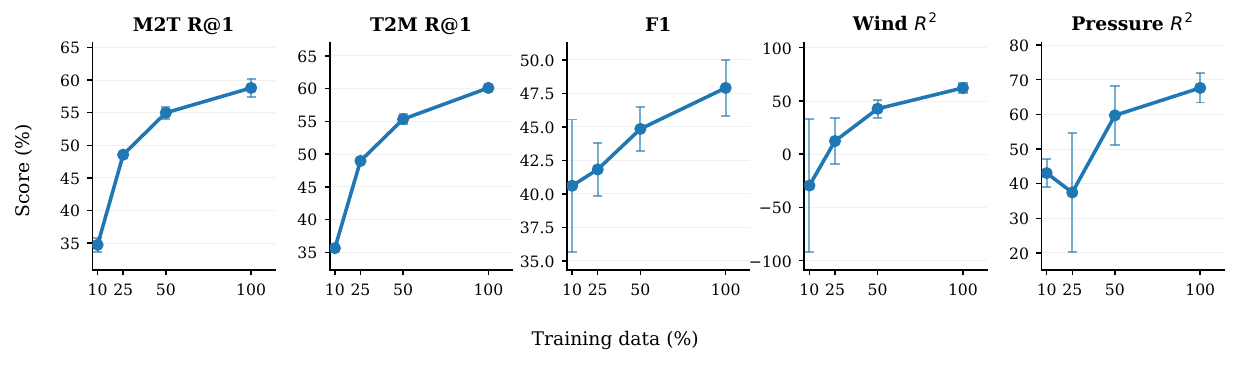}
    \end{center}
    \caption{Effect of training data scale. We train Meteo-CLIP with 10\%, 25\%, 50\%, and 100\% of the training data and evaluate model performance across retrieval, classification, and physical variable estimation. Error bars denote standard deviation over three training runs.}
    \label{fig:training_data_scaling_abl}
\end{figure}

\clearpage

\section{Qualitative Examples}

\begin{figure}[h]
\begin{center}
\includegraphics[width=\textwidth]{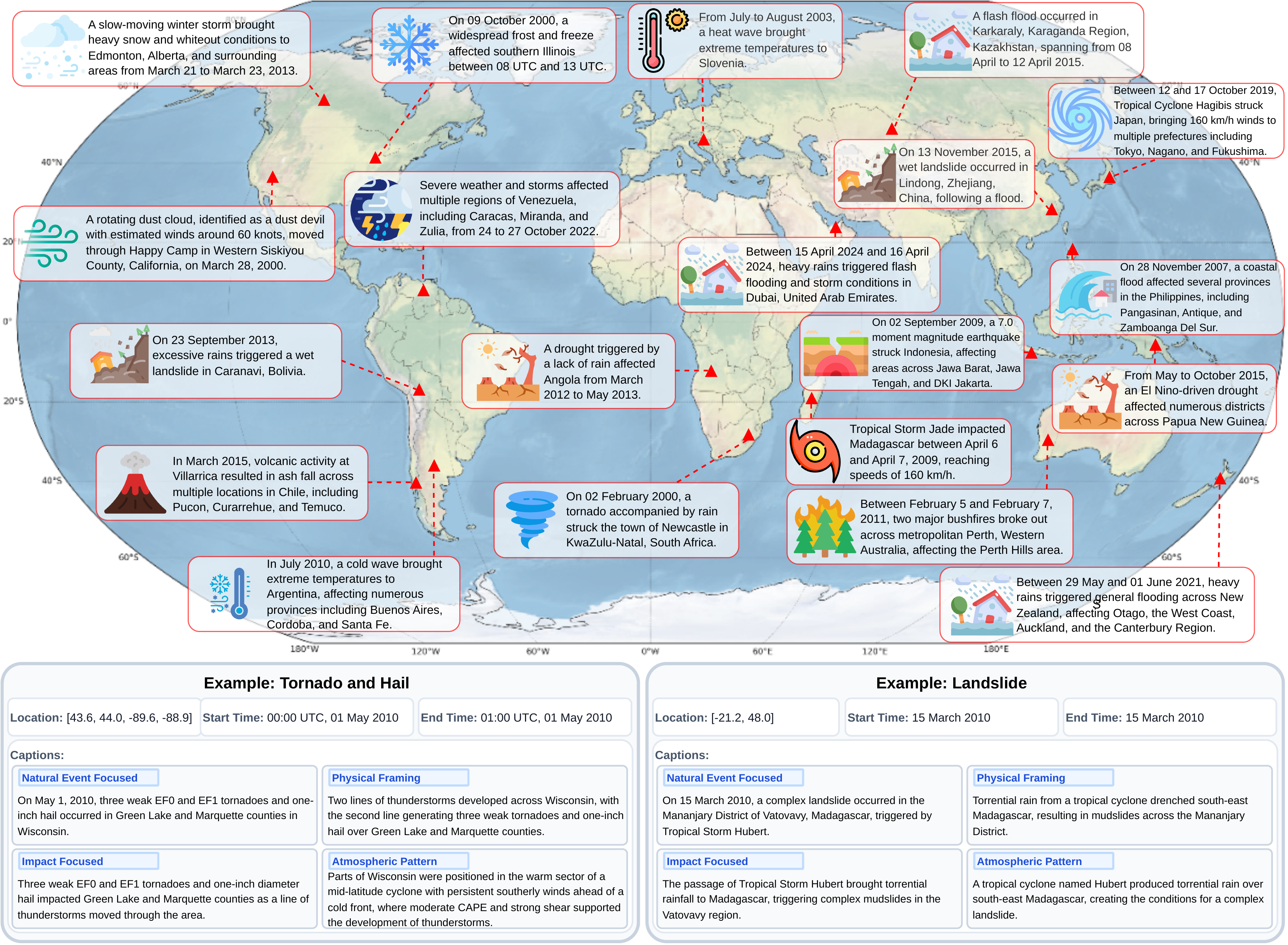}
\end{center}
\caption{Examples from \name{}. The global map illustrates geographically diverse natural-event records, while the bottom panels provide detailed examples of a tornado and hail event, and a landslide with their spatial and temporal metadata along with four grounded caption styles.}
\label{fig:representative_examples_global_map}
\end{figure}

\begin{figure}[h]
    \centering

    \begin{subfigure}[t]{\textwidth}
        \centering
        \includegraphics[width=\textwidth]{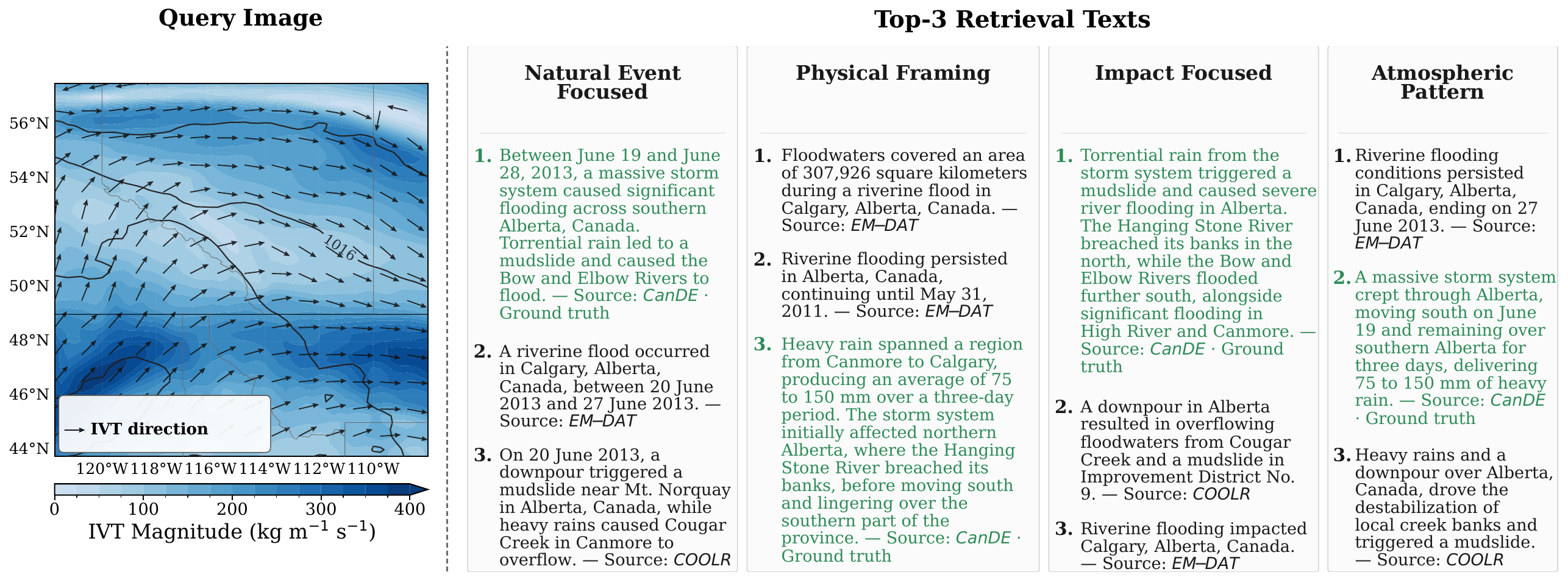}
        \caption{}
    \end{subfigure}

    \begin{subfigure}[t]{0.41\textwidth}
        \centering
        \includegraphics[width=\textwidth]{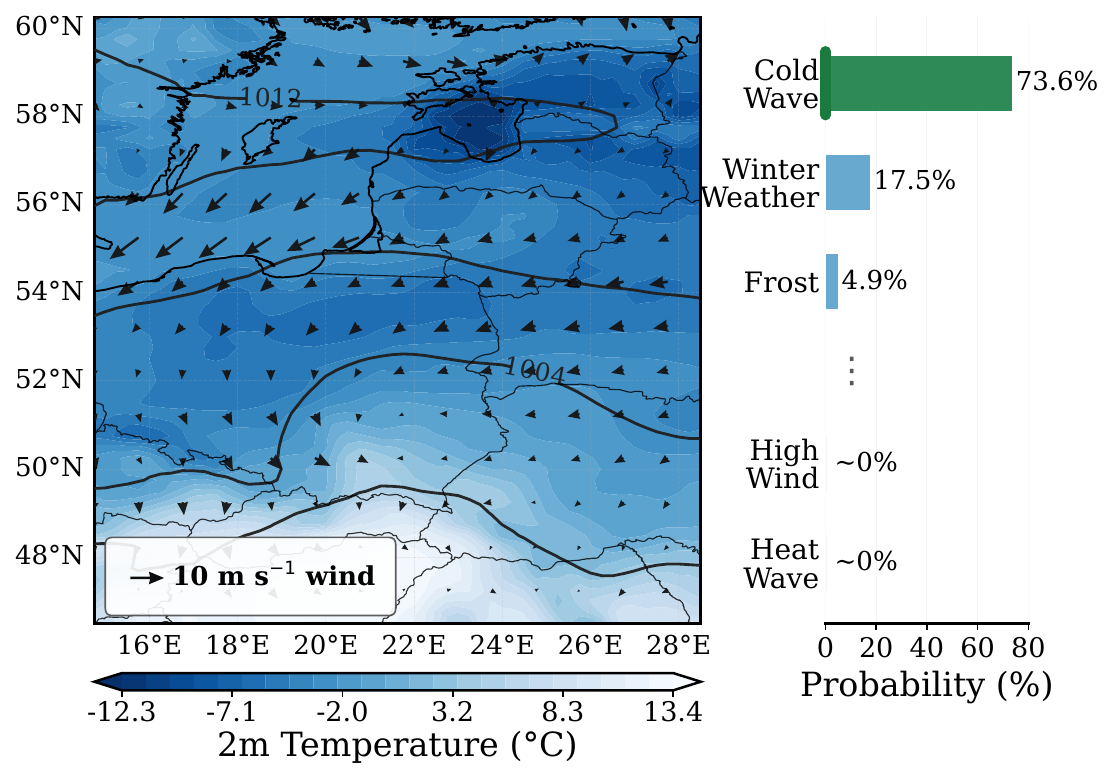}
        \caption{}
    \end{subfigure}
    \hfill
    \begin{subfigure}[t]{0.58\textwidth}
        \centering
        \includegraphics[width=\textwidth]{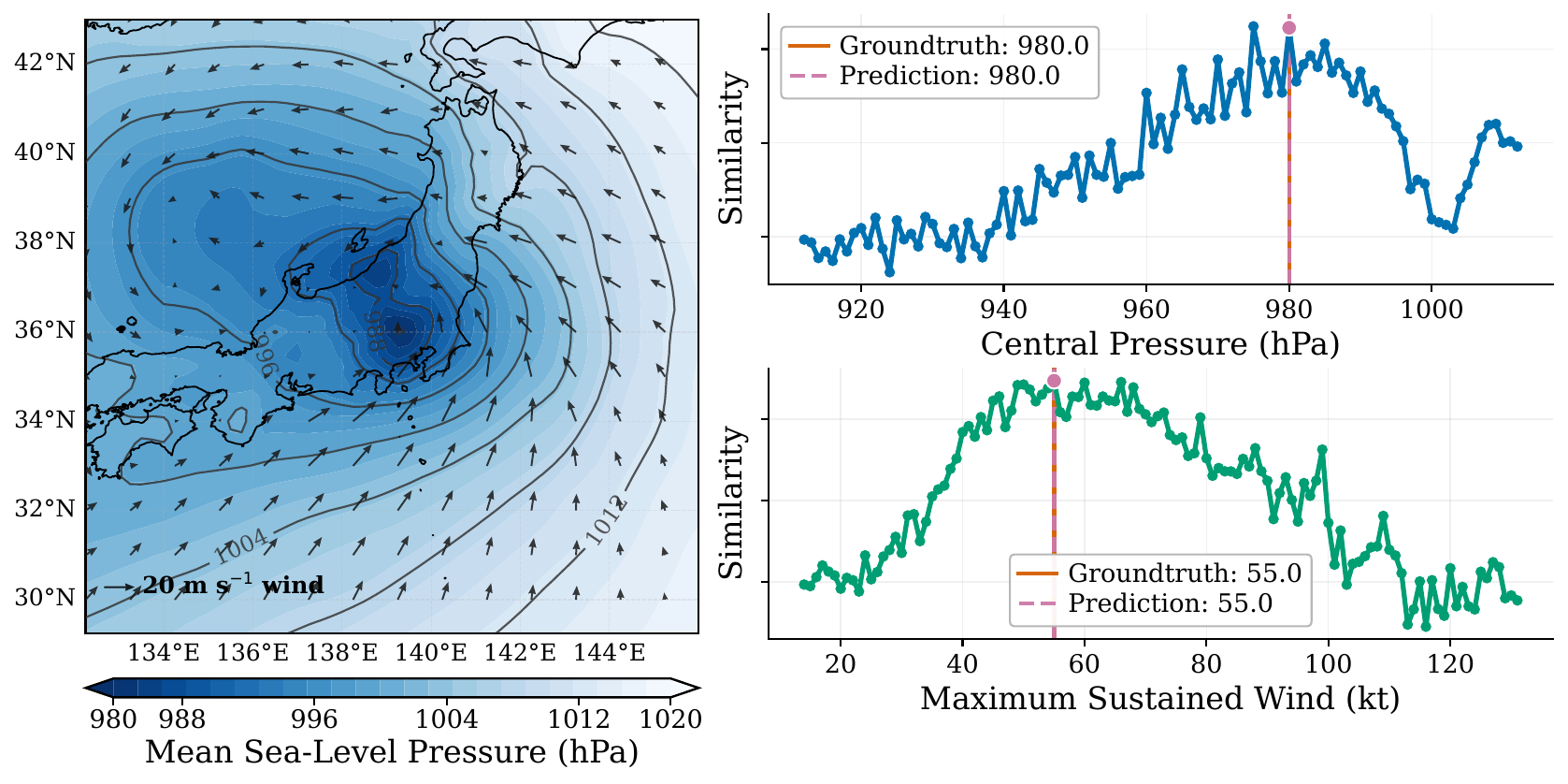}
        \caption{}
    \end{subfigure}

    \caption{\textbf{Qualitative results using the MeteoLIP meteorology and text encoders.} 
(a) M2T retrieval retrieves captions describing the same or closely related event across different caption styles. 
(b) The cold-wave example assigns the highest probability to cold wave, followed by winter weather and frost.
(c) The similarity profiles over candidate physical values show peaks around the predicted minimum central pressure and maximum sustained wind speed.}
    \label{fig:qualitative_results_main}
\end{figure}

\begin{figure}[h]
\begin{center}
\includegraphics[width=\textwidth]{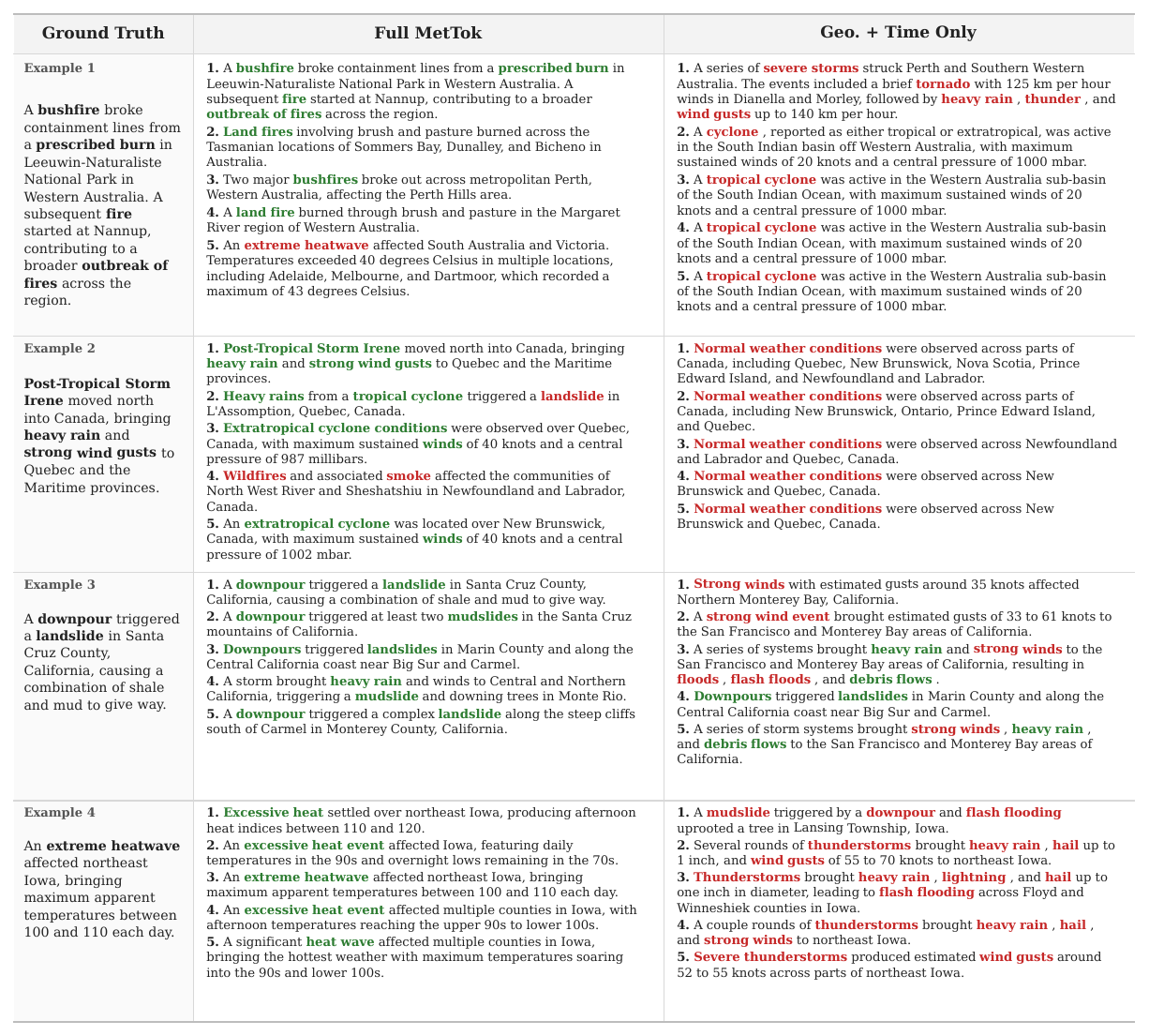}
\end{center}
\caption{Qualitative comparison of M2T retrieval using the full MetTok configuration and geolocation+time only, showing the top-5 retrieved texts for each ground-truth event, with \textcolor{green}{green} indicating meteorologically related events and \textcolor{red}{red} indicating unrelated phenomena. Full MetTok retrieves more meteorologically relevant events, while geolocation and time alone can retrieve unrelated phenomena.}
\label{fig:mettok_qualitative_retrieval}
\end{figure}

\clearpage
\newpage

\section{Dataset Samples Record Construction} \label{app:dataset_samples_record_construction}

\subsection{Expert-Curated Natural Event and Disaster Records} \label{appendix:expert_curated_records}

\begin{table}[t]
\centering

\caption{
Expert-curated natural event and disaster sources used in \name{}.
The table reports the geographic scope, provider type, natural phenomena type, years used, and number of records after source-specific processing.
}

\label{tab:expert_event_sources}

\small
\setlength{\tabcolsep}{3.5pt}
\renewcommand{\arraystretch}{1.08}

\begin{adjustbox}{max width=\linewidth}
\begin{tabular}{@{}llllcr@{}}
\toprule
\textbf{Source}
& \textbf{Geographic Scope}
& \textbf{Provider Type}
& \textbf{Natural Phenomena Type}
& \textbf{Years Used}
& \textbf{Num. of Records} \\
\midrule

EM-DAT
& Global
& Research institution
& Natural hazards and disasters
& 2000--2024
& 25,172 \\

NOAA
& United States
& Government agency
& Severe weather events
& 2000--2024
& 286,292 \\

IBTrACS
& Global
& Government agencies
& Cyclones
& 2000--2024
& 84,764 \\

COOLR
& Global
& Government agency
& Landslides
& 2000--2024
& 12,511 \\

PIKART
& Global
& Research institution
& Atmospheric rivers
& 2000--2023
& 36,840 \\

AusDE
& Australia
& Government agency
& Natural hazards and disasters
& 2000--2014
& 95 \\

MexDE
& Mexico
& Government agency
& Natural hazards and disasters
& 2000--2016
& 1,628 \\

CanDE
& Canada
& Government agency
& Natural hazards and disasters
& 2000--2022
& 385 \\

IndDE
& Indonesia
& Government agency
& Natural hazards and disasters
& 2008--2024
& 34,623 \\

FraDE
& France
& Government agency
& Natural hazards and disasters
& 2000--2024
& 6,875 \\

\bottomrule
\end{tabular}
\end{adjustbox}

\end{table}

We collect natural event and disaster records from 10 expert-curated sources, summarized in Table~\ref{tab:expert_event_sources}. We apply source-specific processing to construct standardized metadata records. The resulting metadata is subsequently used to ground multi-style caption generation. We describe the processing for each source below.

\textbf{NOAA Storm Events Database.}
The event record source \citep{noaa_storm_events} provides records of severe weather events across the United States. We retain records from 2000--2024 and extract the available event metadata, including the episode and event narratives, event type, state and county, start and end times, event magnitude and magnitude type, flood cause, and tornado F/EF-scale rating. We further group event records using the combination of \texttt{EPISODE\_ID} and \texttt{EPISODE\_NARRATIVE}, and aggregate the associated event types, measurements, event narratives, locations, and start and end times. We additionally convert all reported local start and end times to UTC. To obtain a standardized geospatial extent for each episode, we map the reported state and county or weather-zone information to official U.S. geographic boundary data\footnote{\url{https://www2.census.gov/geo/tiger/GENZ2024/shp/cb_2024_us_county_500k.zip},\url{https://www.weather.gov/source/gis/Shapefiles/WSOM/z_18mr25.zip},\url{https://www.weather.gov/source/gis/Shapefiles/WSOM/mz18mr25.zip},\url{https://www.weather.gov/source/gis/Shapefiles/WSOM/oz18mr25.zip},\url{https://www.weather.gov/source/gis/Shapefiles/WSOM/hz20fe25.zip}} using GeoPandas, and convert the matched geographic boundaries to their enclosing latitude--longitude bounding box. We then construct a standardized metadata record containing the episode start and end times, event types, location, available physical measurements, flood cause, tornado F/EF-scale rating, and narrative.

\textbf{IBTrACS.} IBTrACS \citep{ibtracs_paper,ibtracs_data} provides the most comprehensive global archive of tropical cyclones. We retain records from 2000--2024 and extract the available metadata, including basin and sub-basin, cyclone nature, time, center location, WMO wind and pressure, storm translation speed and direction, and agency-provided cyclone classifications. We further retain only main-track records with agency-reported position and intensity values, i.e. those whose \texttt{TRACK\_TYPE} is \texttt{main} and whose \texttt{IFLAG} contains \texttt{O} or \texttt{V}. We exclude records with an unknown or unreported cyclone nature. To obtain more detailed location information, we reverse geocode cyclone center coordinates using OpenStreetMap\footnote{\url{https://www.openstreetmap.org/}} for records whose \texttt{DIST2LAND} is less than 50~km. We additionally derive a more specific cyclone subcategory from the available agency-specific classifications while preserving differences among agency classifications. We then construct a standardized metadata record containing the snapshot time, basin and sub-basin, center location, cyclone nature and subcategory, WMO wind and pressure, and storm translation speed and direction.

\textbf{EM-DAT.} EM-DAT \citep{EM-DAT} provides records of major disasters worldwide. We retain records from 2000--2024 whose disaster group is \texttt{Natural} and whose disaster subgroup is \texttt{Climatological}, \texttt{Meteorological}, \texttt{Hydrological}, or \texttt{Geophysical}, and exclude records with missing start or end years. We extract the available metadata, including disaster type and subtype, event name, country, location, origin, associated disaster types, magnitude and magnitude scale, and start and end dates. Each disaster record is uniquely identified by its \texttt{DisNo.}. To obtain a standardized geospatial extent for each record, we geocode each reported location using OpenStreetMap and aggregate the resulting geospatial boundaries to derive an enclosing latitude--longitude bounding box. For disasters spanning large geographic extents, we further perform spatial upsampling by generating localized bounding boxes for subregions within the original extent. Starting from 9,380 filtered records, our upsampling strategy generates 15,792 additional samples from 2,041 large-extent records. We then construct a standardized metadata record containing the disaster type and subtype, event name, origin, associated disaster types, magnitude and magnitude scale, start and end dates, and locations.

\textbf{COOLR.} COOLR \citep{coolr} provides records of reported landslides and mass movements worldwide. We retain records from 2000--2024 with available latitude and longitude coordinates and extract the available metadata, including event date, event title and description, landslide category and trigger, and coordinates. To obtain a textual location for each event, we reverse geocode the reported latitude--longitude coordinates using OpenStreetMap and exclude records for which reverse geocoding does not return a location. We then construct a standardized metadata record containing the landslide category and trigger, location, event date, event title, and event description.

\textbf{PIKART.} PIKART \citep{pikart} provides a global high-resolution catalog of atmospheric rivers derived from ERA5 meteorological data based on integrated vapor transport (IVT). We retain records from 2000--2023 and select one timestamp from each atmospheric river track. For each selected timestamp, we extract the available metadata, including intersected continents, mean IVT, track centroid speed, and track centroid positions, and use the corresponding atmospheric river extent as the geospatial extent. To provide a qualitative characterization of IVT magnitude, we compute the distribution of mean IVT across all timestamps and tracks and categorize values below the 10th percentile as \texttt{low}, above the 90th percentile as \texttt{high}, and between the 40th and 60th percentiles as \texttt{moderate}. We additionally derive the atmospheric river propagation direction from the displacement of the track centroid between the beginning and end of each track. We then construct a standardized metadata record containing the snapshot time, intersected continents, mean IVT magnitude and category, and track-mean atmospheric river propagation speed and direction.

\textbf{AusDE.} AusDE \citep{ausde} provides records of events and disasters across Australia from a government-curated source. We retain records from 2000--2014, exclude events occurring outside Australia, and remove non-natural events such as transport accidents, criminal acts, and industrial incidents. We extract the available metadata, including event title and description, start and end dates, and latitude--longitude coordinates. To obtain a standardized geospatial extent for each record, we use GPT-5.5 Pro with the event title, description, and provided coordinates to derive a latitude--longitude bounding box corresponding to the affected area. We then construct a standardized metadata record containing the disaster title, start and end dates, and event description.

\textbf{MexDE.} MexDE \citep{mexde} provides records of events and disasters across Mexico from a government-curated source. We retain records from 2000--2024 and exclude non-natural events such as chemical events. We extract the available metadata, including state, municipality, start and end dates, phenomenon type, classification, and observations, and translate the original Spanish-language metadata into English using GPT-5.5. We further group records sharing the same start date, end date, phenomenon type, classification, and observations. To obtain a standardized geospatial extent for each record, we geocode each reported municipality and state using OpenStreetMap, and aggregate the resulting bounding boxes to derive an enclosing latitude--longitude bounding box. We then construct a standardized metadata record containing the phenomenon type, observations, start and end dates, and location.

\textbf{CanDE.} CanDE \citep{cande} provides records of events and disasters across Canada from a government-curated source. We retain records from 2000--2024 whose event group is \texttt{Natural} and exclude records whose event subgroup is \texttt{Biological}. We extract the available metadata, including event group and subgroup, disaster type, start and end dates, reported places, affected provinces, and event comments. To obtain a standardized geospatial extent for each record, we use GPT-5.5 Pro with the reported places, affected provinces, and event comments to derive a latitude--longitude bounding box corresponding to the affected area. We then construct a standardized metadata record containing the disaster type, locations, start and end dates, and event description.

\textbf{IndDE.} IndDE \citep{indde} provides records of events and disasters across Indonesia from a government-curated source. We retain records from 2000--2024 and exclude non-natural incidents such as residential fires, social conflicts, and technological failures. We extract the available metadata, including disaster type, event type, province, regency or city, date, and time, and translate the disaster and event types from Indonesian into English using GPT-5.5. We further group records sharing the same disaster type, event type, province, and date, and aggregate their associated locations. We also remove reported times of \texttt{08:08:08 AM} and \texttt{12:01:00 AM}, which occur anomalously frequently in the source, and convert the remaining local times to UTC. To obtain a geospatial extent for each record, we geocode each reported regency or city using OpenStreetMap, and aggregate the resulting bounding boxes to derive an enclosing latitude--longitude bounding box. We then construct a standardized metadata record containing the disaster type and subtype, date, and locations.

\textbf{FraDE.} FraDE \citep{frade} provides records of events and disasters across France from a government-curated source. We retain records from 2000--2024 and extract the available metadata, including disaster type, start and end dates, and municipality name and code. We further group records sharing the same disaster type, start date, and end date, and aggregate their associated municipalities. We translate the disaster type names from French into English using GPT-5.5. To obtain a standardized geospatial extent for each record, we match each reported municipality to its corresponding French commune\footnote{\href{https://www.data.gouv.fr/datasets/communes-et-villes-de-france-en-csv-excel-json-parquet-et-feather}{French commune boundary data}} and aggregate the resulting administrative boundaries to derive a latitude--longitude bounding box. We then construct a standardized metadata record containing the disaster type, start and end dates, and locations.

\subsection{Climatology-Based Normal-Weather Sampling}
\label{app:normal_weather}

\begin{figure}[t]
    \centering

    \begin{minipage}[c]{0.48\linewidth}
        \centering
        \includegraphics[width=\linewidth]{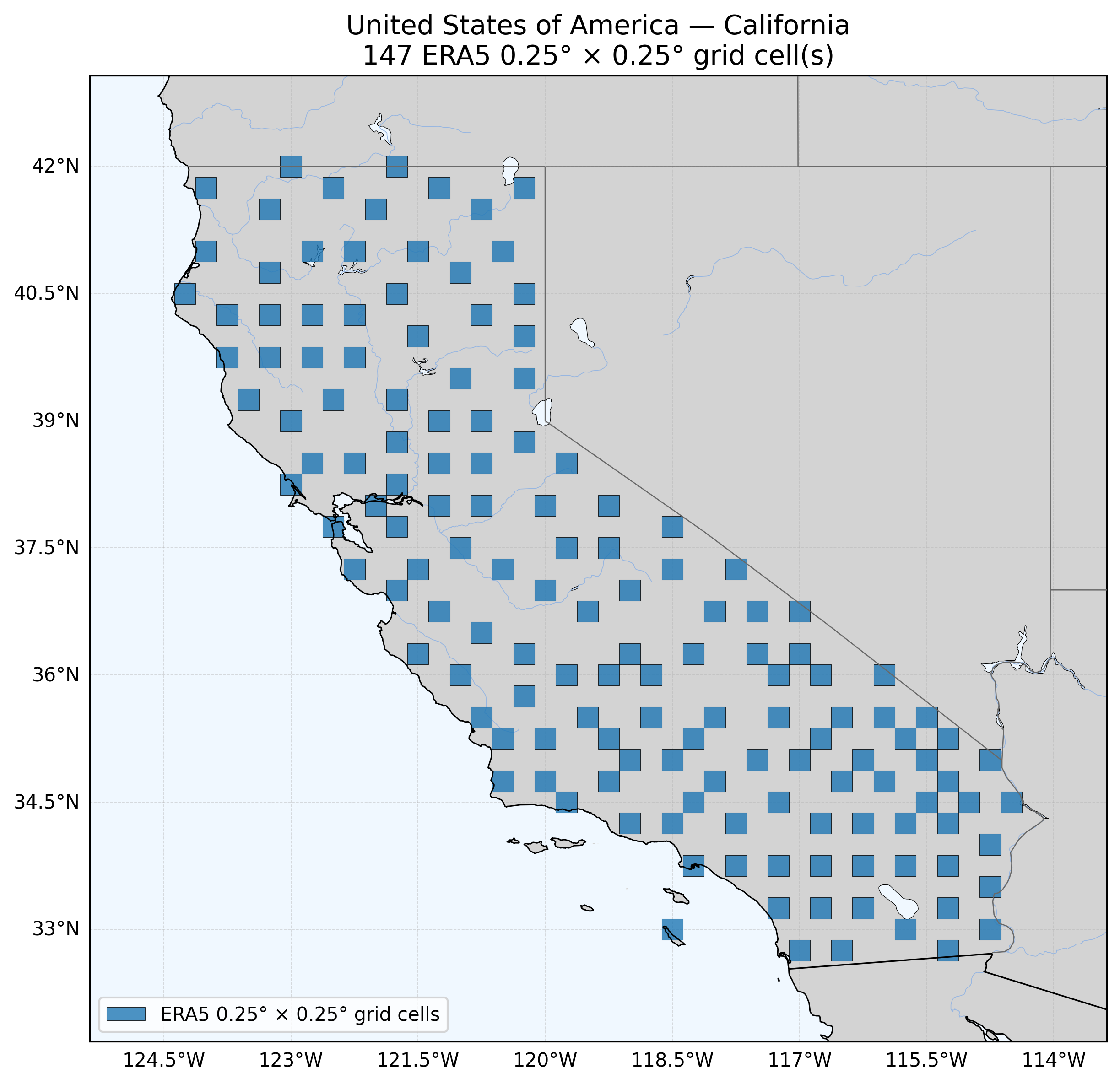}
    \end{minipage}
    \hfill
    \begin{minipage}[c]{0.48\linewidth}
        \centering
        \includegraphics[width=\linewidth]{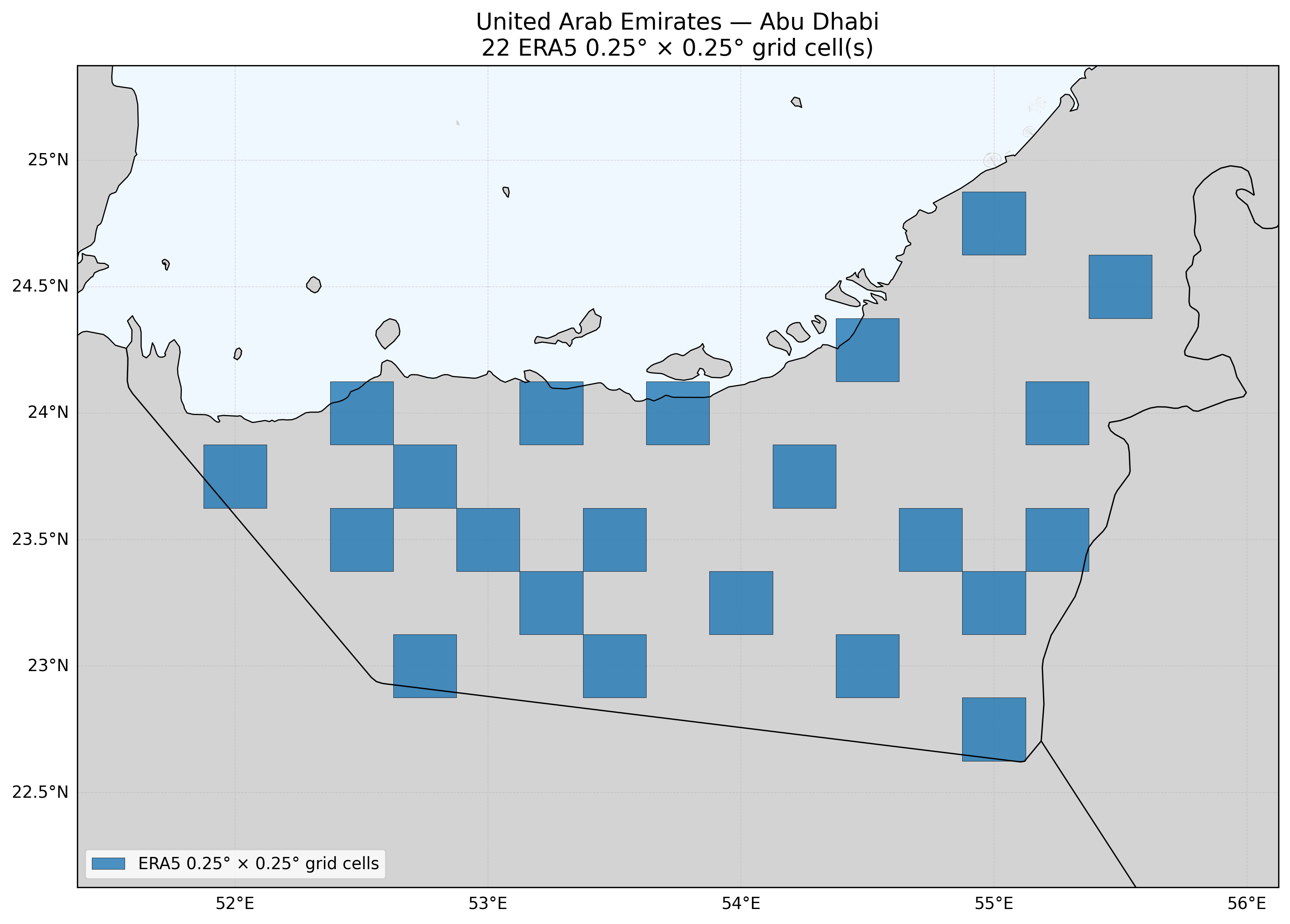}
    \end{minipage}

    \caption{Examples of spatial grid-cell sampling for climatology-based
    normal-weather identification on California and Abu Dhabi.}
    \label{fig:normal_weather_grid_sampling}
\end{figure}

\begin{table}[h]
\centering
\caption{Variable-specific climatological criteria used for normal-weather sampling. Percentiles are computed independently for each ERA5 grid cell, day of year, UTC hour, and variable over the 1991--2020 reference period using a centered $\pm2$-day window.}
\label{tab:normal_weather_thresholds}

\small
\setlength{\tabcolsep}{5pt}
\renewcommand{\arraystretch}{1.08}

\begin{adjustbox}{max width=\linewidth}
\begin{tabular}{@{}lp{0.76\linewidth}@{}}
\toprule
\textbf{Criterion} & \textbf{Variables} \\
\midrule

P05--P95
& 2m temperature; 2m dewpoint temperature; skin temperature; mean sea-level pressure; surface pressure; sea-surface temperature \\

$\leq$ P90
& 10m wind speed; 100m wind speed; 10m wind gust; total precipitation (1h, 3h, 6h, and 24h); 3h mean sea-level pressure change; significant height of combined wind waves and swell; mean wave period \\

$\geq$ P10
& Dewpoint depression; surface solar radiation downwards; cloud base height \\

\bottomrule
\end{tabular}
\end{adjustbox}
\end{table}

For each first-level administrative unit (ADM1), we identify ERA5 grid cells within its geoBoundaries administrative boundary \citep{geoboundaries} and randomly sample 20\% of the cells, with at least 10 cells whenever possible. For instance, as illustrated in Figure~\ref{fig:normal_weather_grid_sampling}, this yields 147 sampled grid cells for California and 22 for Abu Dhabi. For each sampled cell, we retrieve ERA5 hourly variables from 1991 to 2025, including 2m temperature, 2m dewpoint temperature, skin temperature, mean sea-level pressure, surface pressure, sea-surface temperature, 10m and 100m wind speed, 10m wind gust, total precipitation, surface solar radiation downwards, cloud base height, significant height of combined wind waves and swell, and mean wave period, when applicable. We additionally derive 3h, 6h, and 24h precipitation accumulations, 3h mean sea-level pressure change, and dewpoint depression. 
The complete set of variables and their corresponding normality criteria is summarized in Table~\ref{tab:normal_weather_thresholds}. For each grid cell, variable, day of year, and UTC hour, we estimate the climatological thresholds over the 1991--2020 WMO reference period \citep{wmo_clim_30} using observations pooled from the centered $\pm2$-day window, yielding 150 observations for each threshold estimate. We then apply the grid-cell- and ADM1-level normality rules described in Section~\ref{sec:data_construction} to identify eligible normal-weather hours. We further restrict the resulting candidate pool to 2000--2024 and perform global sampling, yielding \numnegsampling{} normal-weather samples.

\begin{figure}[h]
\centering
\includegraphics[width=\linewidth]{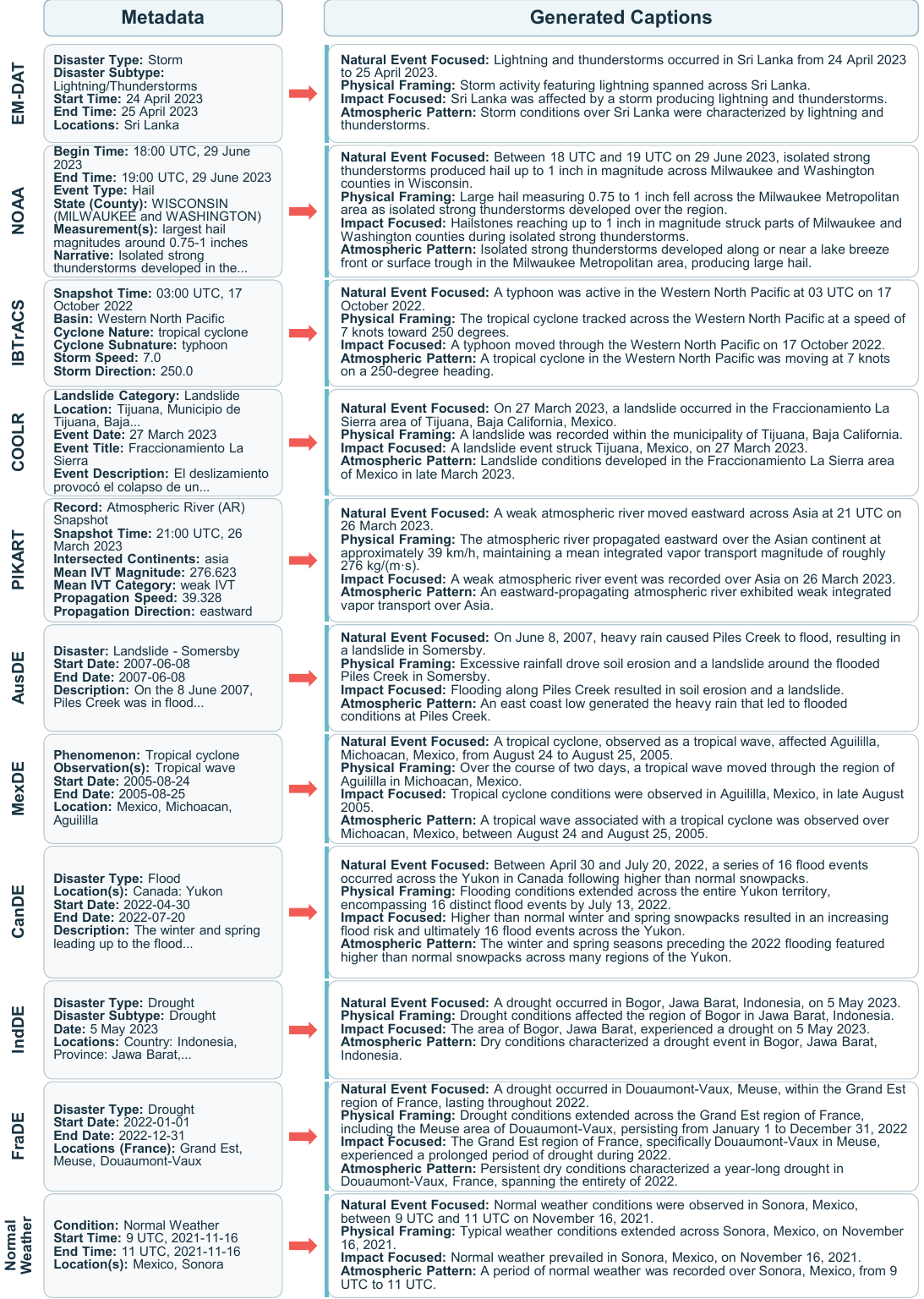}
\caption{Representative metadata records and their generated captions across record sources.}
\label{fig:caption_generation_examples}
\end{figure}

\section{Multi-Style Grounded Caption Generation and Audit} \label{app:caption_generation_and_audit}

\subsection{Caption Generation}
\label{app:caption_generation}

\begin{figure}[h]
\centering
\includegraphics[width=\linewidth]{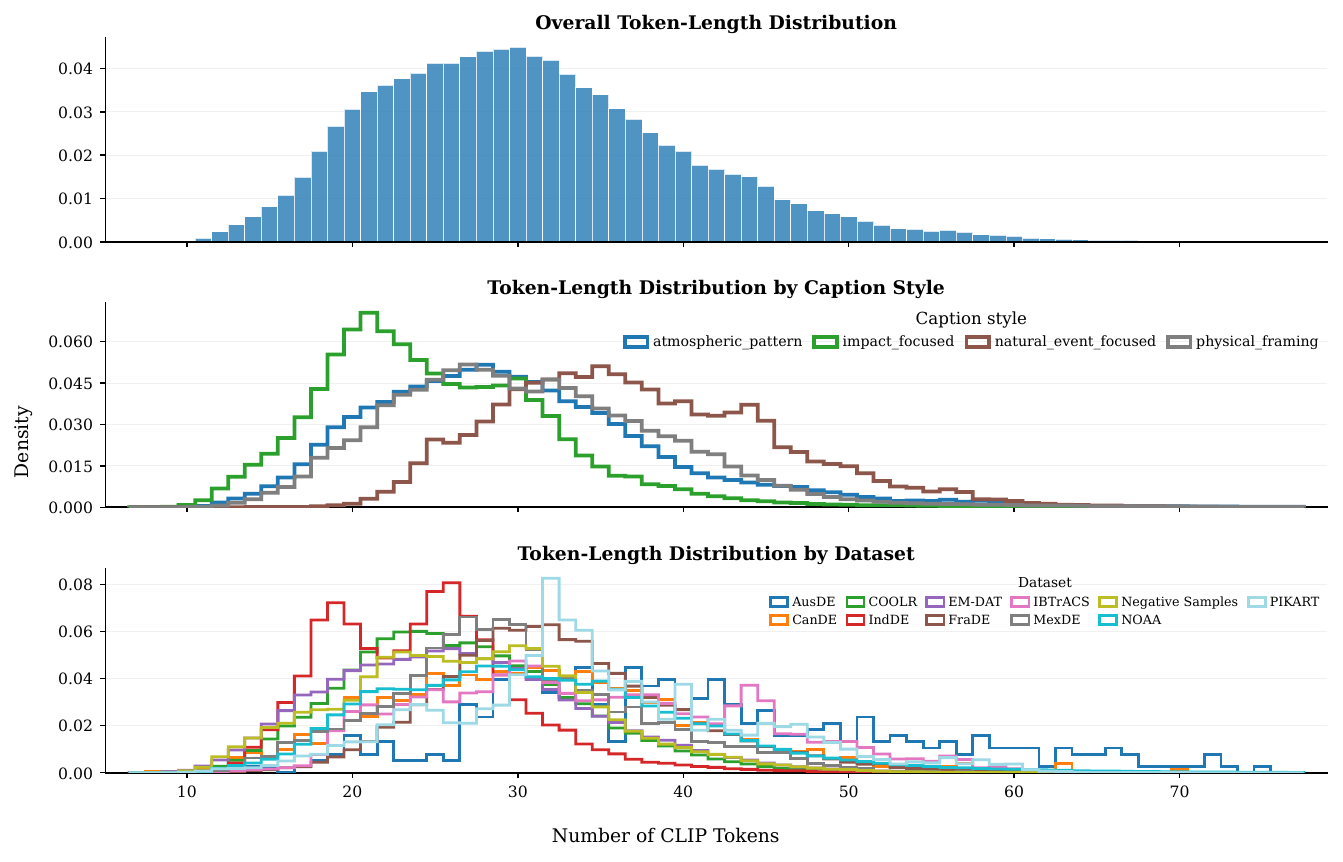}
\caption{Distribution of generated caption lengths measured by using the CLIP text tokenizer. We report the overall distribution, distributions across the four caption styles, and distributions across record data sources.}
\label{fig:caption_length_distribution}
\end{figure}

For each metadata record, we use Gemini 3.1 Pro to generate one caption in each of four styles. \textit{Natural-event-focused} captions describe the event or phenomenon directly, with time and location information when available; \textit{physical-framing} captions emphasize its supported spatial or temporal organization; \textit{impact-focused} captions emphasize supported weather-, climate-, environmental-, or hazard-related effects; and \textit{atmospheric-pattern} captions emphasize provided meteorological or environmental conditions and setup. The generation prompt requires all captions to be directly supported by the input metadata and prohibits unsupported causes, impacts, forecasts, and descriptive modifiers. When the metadata does not sufficiently support a requested style, the model is instructed to produce a minimal grounded caption rather than infer missing style-specific content. The representative metadata and generated captions are shown in {Figure~\ref{fig:caption_generation_examples}}. We further report the distribution of caption lengths in Figure~\ref{fig:caption_length_distribution}. The full generation prompt is provided below.

\begin{promptbox}{Grounded Caption Generation Prompt}
\label{prompt:caption_generation}
\footnotesize
You are an expert caption writer for climate and weather data.

\textbf{Task:} Generate grounded captions from the provided metadata/context.

\textbf{Goal:} Produce stylistically diverse captions that are strictly supported by the input, preserve informative uncertainty when present, and are suitable for text-ERA5 climate data alignment.

\textbf{Hard constraints:} \begin{itemize}[leftmargin=*, itemsep=1pt, topsep=2pt, parsep=0pt, partopsep=0pt]
\item Do not add any fact that is not directly supported by the input.
\item Do not infer specific causes, future outcomes, identities, or impacts unless explicitly given.
\item Do not describe objects, actions, regions, or conditions not present in the metadata/context.
\item If a detail is missing, leave it out.
\item If the input contains explicit uncertainty, disagreement, multiple reported classifications, or alternative labels, do not replace them with a single definite claim. Retain the ambiguity when it is relevant to the caption.
\item Do not mention missing, unspecified, or unavailable information.
\item Do not force all supported metadata into every caption. Include only the details that best fit the requested style.
\item Do not refer to the metadata, narrative, source text, or report inside the caption.
\item Do not convert numeric or categorical metadata into qualitative descriptions unless that wording is directly supported by the input.
\item Do not add unsupported descriptive modifiers such as compact, broad, organized, intense, weak, rapidly, slowly, strengthening, or weakening.
\item Do not mention vehicles, casualties, injuries, deaths, missing people, houses, buildings, infrastructure, property damage, economic losses, or other human-impact or built-environment details. Focus only on the supported natural phenomena, weather conditions, environmental conditions, or natural hazard type.
\item If a requested style is not supported, still produce that caption, but reduce it to a minimal, fully grounded restatement of the event rather than inventing missing structure, impacts, or atmospheric setup.
\item Do not make every caption a full recap of the same event metadata.
\item When mentioning time, use hour-only format such as 21 UTC, not 21:00 UTC.
\item Do not repeat the same system name, event name, storm name, phenomenon name, or cyclone name in every caption.
\item Do not mention exact time in every caption. Use time only when it is important to the grounded description or helps distinguish the event more clearly.
\end{itemize}

\textbf{Writing requirements:}\begin{itemize}[leftmargin=*, itemsep=1pt, topsep=2pt, parsep=0pt, partopsep=0pt]
\item Write exactly one caption for each style.
\item Keep each caption natural, clear, and concise.
\item Ensure each caption reads smoothly as a standalone caption.
\item Each caption must stand on its own and should not assume prior context.
\item Avoid generic referents such as "the system" when a more natural grounded alternative can be used.
\item Prefer caption-like phrasing over report-like phrasing.
\item Make the captions stylistically different by varying emphasis, structure, and phrasing, not by introducing unsupported facts or unsupported interpretations.
\item Avoid repeating the same wording, sentence structure, or lead phrase across captions.
\item Prefer a single sentence, but use 2 to 3 sentences when needed to capture supported details.
\item Correct obvious surface errors such as typos, stray punctuation, spacing issues, duplicated words, and minor formatting problems when the intended meaning is clear.
\item Write all location names in plain ASCII. Use common English names or ASCII romanized forms for locations that contain accents, non-Latin scripts, or special punctuation.
\item Do not discard informative input details solely because they are uncertain or reported differently; maintain their uncertainty status when relevant to the caption.
\item If the input includes too many location names, you may summarize them with a shorter shared parent location only when that summary is clearly supported by the input and does not introduce a new regional interpretation.
\item If you mention a date, you must include the full supported date including the year.
\item Before writing each caption, select only the input details most relevant to that style, and do not reuse the same full set of details across all four captions unless the input does not provide enough distinct supported details to differentiate them further.
\item When the input lacks enough distinct supported details for strong stylistic separation, keep weakly supported captions shorter and more selective rather than repeating the same full metadata summary.
\item Express numeric metadata naturally, and avoid raw technical notation when it does not read smoothly as a caption.
\end{itemize}

\textbf{Allowed caption styles:} \begin{enumerate}[leftmargin=*, itemsep=1pt, topsep=2pt, parsep=0pt, partopsep=0pt]
\item natural\_event\_focused \begin{itemize}[leftmargin=*, itemsep=1pt, topsep=2pt, parsep=0pt, partopsep=0pt]
   \item Focus on what is happening in a straightforward and natural way.
   \item Do not include unrelated human-interest details, casualty counts, or incidental impacts.
   \item Date and time may be included when it helps anchor the event naturally.
   \end{itemize}

\item physical\_framing \begin{itemize}[leftmargin=*, itemsep=1pt, topsep=2pt, parsep=0pt, partopsep=0pt]
   \item Focus on how the event or conditions were distributed, organized, extended, or expressed across space or time.
   \item Prefer wording about extent, spread, concentration, coverage, or visible physical conditions when supported.
   \item Do not simply restate the natural-event caption with reordered wording.
   \item Do not include time unless it is needed to describe the supported spatial or temporal arrangement.
   \end{itemize}

\item impact\_focused \begin{itemize}[leftmargin=*, itemsep=1pt, topsep=2pt, parsep=0pt, partopsep=0pt]
   \item Focus on supported climate-, weather-, environmental, or hazard-related effects of the event.
   \item Avoid named individuals, casualty counts, and highly itemized damage unless explicitly needed.
   \item Do not center unrelated human-interest or incidental impacts.
   \item If only human-loss impacts are supported, describe the event itself without adding consequences.
   \item Do not include time unless it is needed to anchor the reported impacts.
   \end{itemize}

\item atmospheric\_pattern \begin{itemize}[leftmargin=*, itemsep=1pt, topsep=2pt, parsep=0pt, partopsep=0pt]
   \item Focus on atmospheric, meteorological, or environmental setup only when it is explicitly described in the input.
   \item Do not introduce meteorological structure unless it is supported.
   \item If no such pattern is explicitly supported, keep the caption grounded by describing the reported conditions.
   \item Do not include time unless it is needed to ground the reported setup or evolution.
   \end{itemize}
\end{enumerate}

\textbf{Final check before answering:} \begin{itemize}[leftmargin=*, itemsep=1pt, topsep=2pt, parsep=0pt, partopsep=0pt]
\item Is every statement directly supported by the input?
\item Are unsupported causes, impacts, and forecasts excluded?
\item Were obvious surface errors cleaned up without changing the meaning or introducing unsupported interpretations?
\item If multiple locations were summarized, is the shared location label clearly supported by the input?
\item Is each caption aligned with its style?
\item Is each caption concise and natural?
\item Is the output valid JSON with exactly the required keys?
\end{itemize}

\textbf{Output format:} \\
Return valid JSON only, with exactly the required keys and string values. \\
Do not output markdown, comments, code fences, or any text outside the JSON object.

\textbf{Input metadata/context:} \\
\texttt{\{input\_placeholder\}}

\end{promptbox}

\subsection{Caption Audit and Error Analysis}
\label{app:caption_audit}

\textbf{Caption audit.}
A trained graduate annotator manually audits 440 captions,
{comprising 10 records from each of 11 record sources (i.e. 10 expert-curated natural event and disaster sources and the normal-weather samples) across all four caption styles.}
Each caption is evaluated for \textit{date/time accuracy}, \textit{location accuracy}, \textit{event/phenomenon accuracy}, \textit{trigger/cause accuracy}, \textit{unsupported added detail}, \textit{unsupported modifiers}, \textit{weather/climate focus}, and \textit{standalone clarity}. Table~\ref{tab:caption_audit} summarizes the results, with a caption counted as passing only when all audit criteria are satisfied.

\textbf{Error analysis.} Among the erroneous captions, the most frequent audit failure category is weather/climate focus (7 cases), followed by location accuracy (6), unsupported modifiers (2), trigger/cause accuracy (2), unsupported added detail (1), and event/phenomenon accuracy (1). Table~\ref{tab:caption_failure_examples} provides representative examples. In particular, several impact-focused failures emphasize human consequences such as drowning or transport accidents rather than focusing on weather, climate, environmental, or hazard-related effects. Other errors arise from overgeneralized locations, unsupported qualitative modifiers, or descriptions that alter the identity of the reported phenomenon.

\begin{table}[h]
\centering
\caption{Manual audit results for generated captions. A trained graduate annotator verified 10 samples from each record source across all caption styles, totaling 440 captions.}
\label{tab:caption_audit}

\small
\setlength{\tabcolsep}{5pt}
\renewcommand{\arraystretch}{1.08}

\begin{tabular}{@{}lrrr@{}}
\toprule
\textbf{Caption Style}
& \textbf{Audited}
& \textbf{Passed}
& \textbf{Pass Rate} \\
\midrule
Natural Event Focused & 110 & 110 & 100.00\% \\
Physical Framing      & 110 & 108 & 98.18\% \\
Impact Focused        & 110 & 98  & 89.09\% \\
Atmospheric Pattern   & 110 & 105 & 95.45\% \\
\midrule
\textbf{Overall} & \textbf{440} & \textbf{421} & \textbf{95.68\%} \\
\bottomrule
\end{tabular}
\end{table}

\begin{table}[t]
\centering
\caption{Representative caption audit failures and their corresponding failure categories.}
\label{tab:caption_failure_examples}

\small
\setlength{\tabcolsep}{4pt}
\renewcommand{\arraystretch}{1.10}

\begin{adjustbox}{max width=\linewidth}
\begin{tabular}{
@{}
>{\raggedright\arraybackslash}p{0.08\linewidth}
>{\raggedright\arraybackslash}p{0.44\linewidth}
>{\raggedright\arraybackslash}p{0.44\linewidth}
@{}}
\toprule
\textbf{Style}
& \textbf{Generated Caption}
& \textbf{Audit Comment} \\
\midrule

Impact Focused
& A rip current resulted in a {\color{red}drowning} in Coastal Horry, South Carolina, on 13 June 2022.
& \textit{Weather/climate focus:} drowning is a human-impact outcome rather than a weather-, climate-, or environmental effect. \\

Impact Focused
& A {\color{red}weak IVT event} carried moisture at a mean magnitude of 249.224~kg/(m$\cdot$s).
& \textit{Event/phenomenon accuracy:} a weak IVT event does not necessarily correspond to an atmospheric river and changes the identity of the reported phenomenon. \\

Impact Focused
& Hazardous snow and ice accumulations from Storm Bettina led to {\color{red}transport accidents} in German locations like Schwabisch Hall and Denzeheide.
& \textit{Weather/climate focus:} transport accidents are incidental human impacts and should not be the focus of the caption. \\

Impact Focused
& The winter storm produced {\color{red}extreme} snow drifts in Prince Edward County, alongside significant snow accumulation and drifting along the north shore of Lake Ontario.
& \textit{Unsupported modifier:} ``extreme'' is not supported by the input metadata. \\

Physical Framing
& Significant precipitation and {\color{red}lake effect snow spread across Eastern Canada}, with {\color{red}intense} snow drifting extending along the north shore of Lake Ontario through to Kingston.
& \textit{Location/modifier:} the lake-effect snow is described more specifically along Lake Ontario through Kingston, thus overgeneralizing its spatial extent; \textit{Unsupported modifier:} ``intense'' is unsupported. \\

\bottomrule
\end{tabular}
\end{adjustbox}

\end{table}

\section{ERA5 Data Alignment and Preprocessing}
\label{app:era5_variables}

\textbf{ERA5 variables.}
We retrieve ERA5 hourly data on single and pressure levels through the Copernicus Climate Data Store (CDS) API. Each meteorological input contains 28 single-level variables, comprising 25 atmospheric and land variables and 3 wave variables, together with 7 pressure-level variables at 925, 850, 700, 500, and 300~hPa. This results in 63 variables in total. The complete variable set is listed in Table~\ref{tab:era5_variables}.

\begin{table}[t]
\centering
\caption{ERA5 variables used to construct the meteorological inputs. The 28 single-level variables are combined with 7 pressure-level variables at 5 pressure levels, yielding 63 variables in total.}
\label{tab:era5_variables}

\small
\setlength{\tabcolsep}{4pt}
\renewcommand{\arraystretch}{1.08}

\begin{adjustbox}{max width=\linewidth}
\begin{tabular}{@{}p{0.23\linewidth}p{0.73\linewidth}@{}}
\toprule
\textbf{Variable Group}
& \textbf{Variables} \\
\midrule

Single-level atmospheric and land
& 2m temperature; 2m dewpoint temperature; 10m u-component of wind; 10m v-component of wind; instantaneous 10m wind gust; mean sea-level pressure; total precipitation; convective precipitation; snowfall; snow depth; snowmelt; runoff; volumetric soil water layer 1; volumetric soil water layer 2; volumetric soil water layer 3; volumetric soil water layer 4; convective available potential energy; convective inhibition; boundary layer height; cloud base height; total cloud cover; zero-degree level; total column water vapour; vertical integral of eastward water vapour flux; vertical integral of northward water vapour flux \\

Single-level wave
& Significant height of combined wind waves and swell; mean wave direction; peak wave period \\

Pressure-level
& Geopotential; temperature; u-component of wind; v-component of wind; specific humidity; vertical velocity; vorticity, each at 925, 850, 700, 500, and 300~hPa \\

\bottomrule
\end{tabular}
\end{adjustbox}

\end{table}

\textbf{Temporal alignment.}
We align each processed record to a single ERA5 hour based on its available temporal information. For IBTrACS, PIKART, and NOAA or IndDE records with an exact timestamp, we use the reported timestamp directly. For NOAA and climatology-based normal-weather records specified over a start--end time interval, we uniformly sample one hourly timestamp within the reported interval. COOLR and some IndDE records provide only a date; for these records, we uniformly sample one of the 24 hours within the reported day in local time and subsequently convert it to UTC. EM-DAT, FraDE, CanDE, AusDE, and MexDE provide date ranges, from which we uniformly sample one hourly timestamp within the reported interval. All reported or sampled timestamps are standardized to UTC before ERA5 retrieval.

\textbf{Spatial alignment.}
The processed records contain either a spatial bounding-box extent or a single center location. NOAA, EM-DAT, PIKART, AusDE, MexDE, CanDE, IndDE, FraDE, and climatology-based normal-weather records provide bounding-box extents, whereas IBTrACS and COOLR provide only a center latitude--longitude coordinate. For IBTrACS and COOLR, we construct a $14^\circ\times14^\circ$ region centered on the reported coordinate. For records with an existing bounding box smaller than $14^\circ\times14^\circ$, we preserve its center and symmetrically expand each dimension to a minimum extent of $14^\circ$ whenever possible. Near the northern or southern geographic boundary, where symmetric expansion would extend beyond $90^\circ$ or $-90^\circ$ latitude, we shift the excess latitudinal extent in the opposite direction. For records whose extent exceeds $14^\circ$ in either dimension, we use the larger of the latitudinal and longitudinal extents and symmetrically expand the shorter dimension to construct a square region containing the original bounding box. 

\textbf{Standardization.}
We retrieve ERA5 patches based on the selected timestamp and spatial extent. We then standardize each extracted patch to $56\times56$ grid cells using linear interpolation for all variables except mean wave direction, for which we use nearest-neighbor interpolation. We further normalize each variable using its mean and
standard deviation computed across samples.

{
\section{Additional Dataset Statistics and Distributions} \label{app:data_statistics_and_distributions}
\textbf{Regional distribution.}
Figure~\ref{fig:regional_distribution} shows the geographic distribution of \name{} across 19 regions. As illustrated, \name{} spans a broad range of geographic regions with North America being the most represented, largely reflecting the NOAA Storm Events Database which contributes 286,292 records from the United States. The remaining global and country-specific sources provide additional coverage across other regions, resulting in broad geographic coverage with a notable concentration in North America.

\textbf{Weather condition class distribution.}
Figure~\ref{fig:classification_train_test_distribution} shows the class distribution for the weather-condition classification task across the training and test sets. The two splits exhibit broadly similar distributions across most classes, with drought showing the largest relative difference. The resulting class imbalance reflects the underlying distribution of records across the constituent record sources.

\begin{figure}[t]
    \centering
    \includegraphics[width=\textwidth]{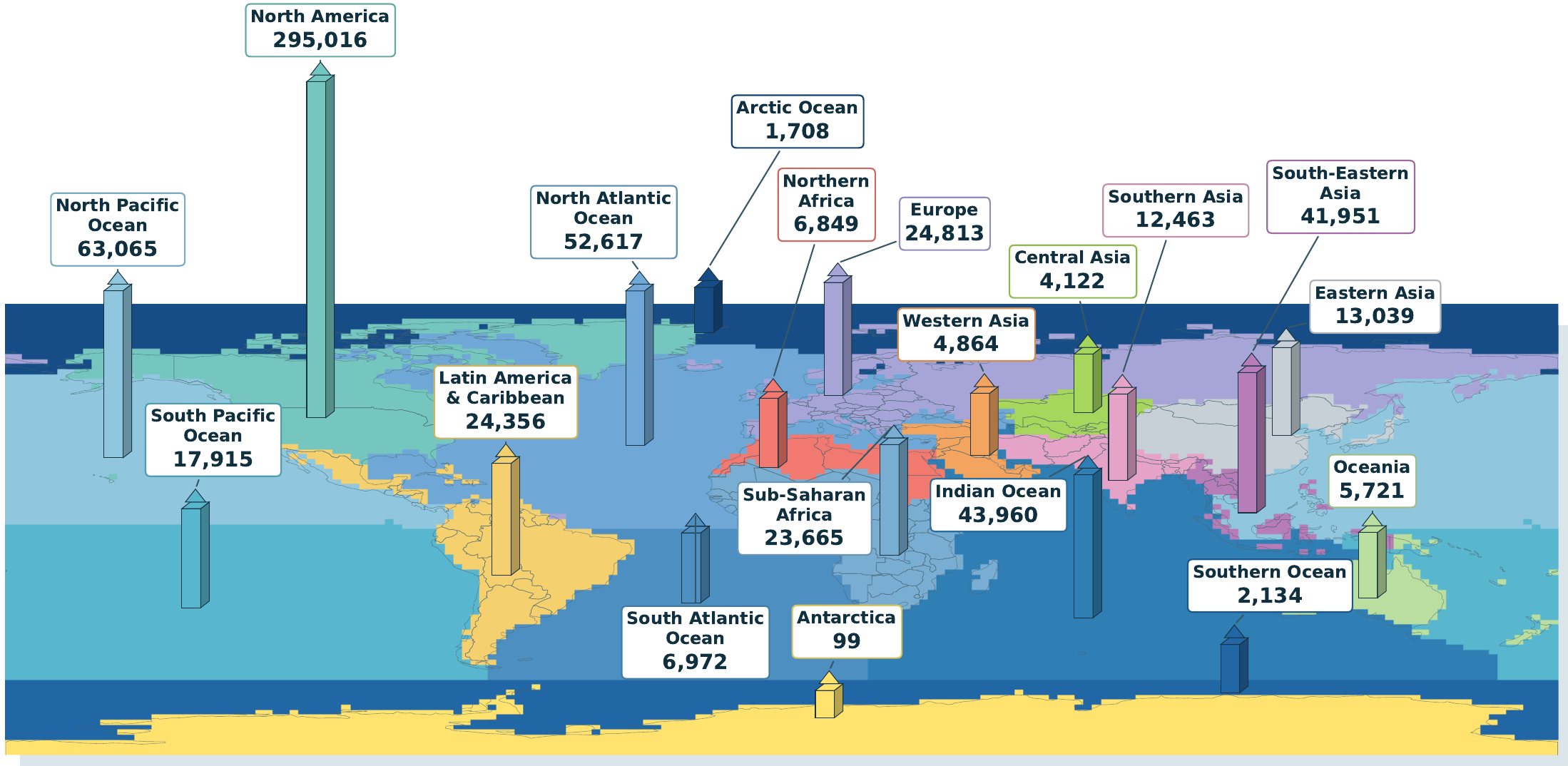}
    \caption{Regional distribution of \name{} records. Records spanning multiple regions contribute to more than one regional count.}
    \label{fig:regional_distribution}
\end{figure}
\begin{figure}[t]
    \centering
    \includegraphics[width=\textwidth]{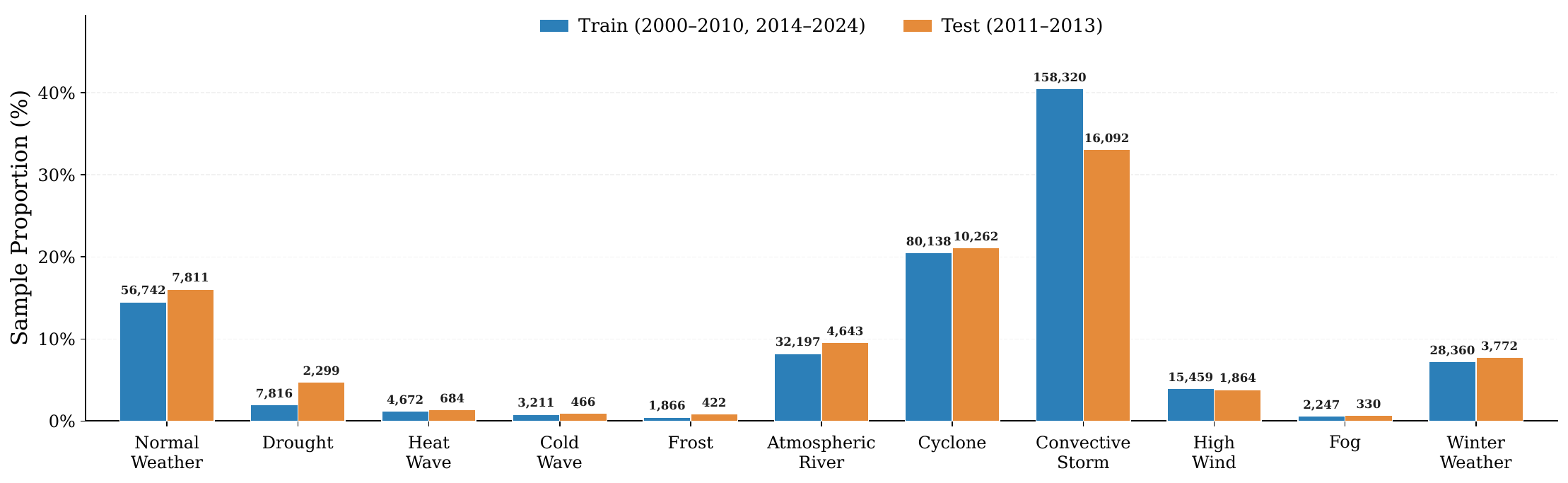}
    \caption{Class distribution for the weather condition classification task in the training and test sets.}    \label{fig:classification_train_test_distribution}
\end{figure}
}

\section{Additional Analysis}

\subsection{Weather Condition Classification Analysis}
\label{app:classification_category_result}

\begin{figure}[t]
    \centering
    \includegraphics[width=\textwidth]{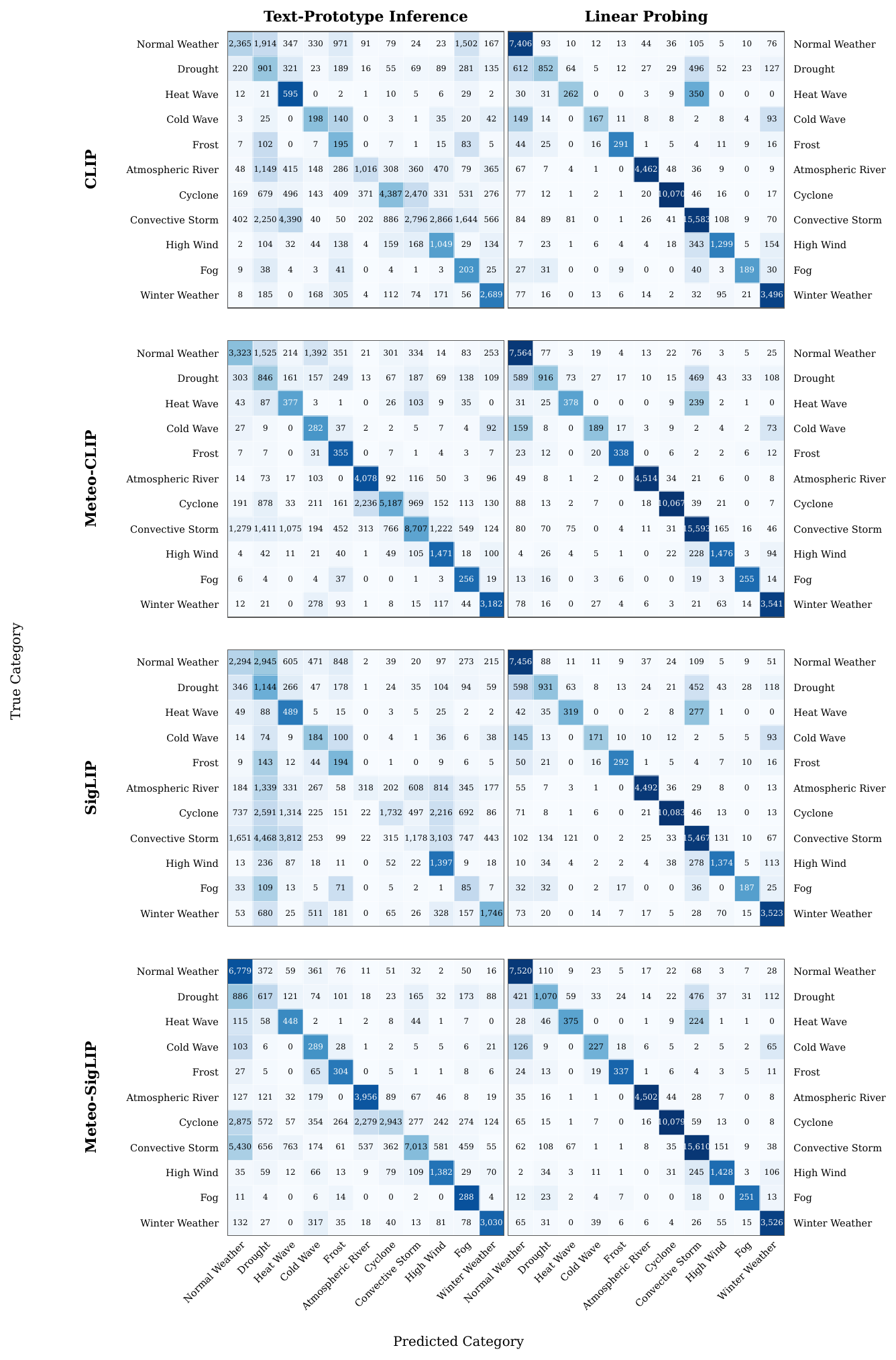}
    \caption{Confusion matrices under \textit{text-prototype inference} and \textit{linear probing}. Cell values report mean counts, while color intensity is normalized by true category to account for class imbalance.}
    \label{fig:confusion_matrix}
\end{figure}

Table~\ref{tab:classification_results_by_category} provides category-wise classification results, with the corresponding confusion matrices shown in Figure~\ref{fig:confusion_matrix}. Under text-prototype inference, Meteo-CLIP and Meteo-SigLIP generally improve category-level performance over their respective CLIP and SigLIP baselines. The improvements are particularly notable for atmospheric river, convective storm, high wind, and winter weather. Under linear probing, performance increases substantially across almost all categories, with Meteo-CLIP and Meteo-SigLIP achieving macro F1 scores of 79.98 and 80.75 respectively. The consistently stronger linear-probing results indicate that the frozen meteorology representations encode substantial category-level information.

Across weather conditions, drought, heat wave, and cold wave remain the most challenging categories. Even under linear probing, their F1 scores are substantially below those of atmospheric river, cyclone, and convective storm, which exceed 94 F1. The lower performance of drought, heat wave, and cold wave cannot be solely attributed to limited test data, as frost achieves above 82 F1 with both MeteoLIP variants despite having only 422 test samples, fewer than drought, heat wave, and cold wave. The confusion matrices further show that these challenging categories are frequently confused with other meteorological conditions, including normal weather.

The lower performance on drought, heat wave, and cold wave potentially reflects their inherently temporal characteristics. For instance, the Copernicus Warm and Cold Spell indicators define such events from sequences of at least three hot or cold days based on daily minimum and maximum temperature anomalies\footnote{\href{https://drought.emergency.copernicus.eu/data/factsheets/factsheet_warmColdSpellIndex.pdf}{EDO INDICATOR FACTSHEET  - Warm Spell and Cold Spell Duration and Intensity Indicators}}. Similarly, meteorological drought indicators such as the Standardized Precipitation Index (SPI) characterize precipitation deficits over accumulated time scales ranging from months to years\footnote{\href{https://drought.emergency.copernicus.eu/data/factsheets/factsheet_spi.pdf}{EDO AND GDO INDICATOR FACTSHEET - Standardized Precipitation Index (SPI)}}. In contrast, MeteoLIP currently takes a single meteorological snapshot as input, thereby limiting its ability to capture the temporal persistence and accumulated anomalies underlying these conditions. These findings motivate future extensions of MeteoLIP to incorporate multiple timestamps or temporal sequences, which can help analyze meteorological conditions over time.

\begin{table}[t]
\caption{Per-category classification performance under \textit{text-prototype inference} and \textit{linear probing} across 11 weather conditions. We report class-wise F1 scores and macro F1, with the number of test samples shown in parentheses. \textbf{Bold} indicates the best result and \underline{underline} indicates the second-best.}
\label{tab:classification_results_by_category}
\centering
\small
\setlength{\tabcolsep}{4pt}
\renewcommand{\arraystretch}{1.05}

\begin{adjustbox}{max width=\linewidth}
\begin{tabular}{@{}clcccc@{}}
\toprule
 & \textbf{Category} & \textbf{CLIP} & \textbf{Meteo-CLIP} & \textbf{SigLIP} & \textbf{Meteo-SigLIP} \\
\midrule
\multirow{12}{*}{\rotatebox[origin=c]{90}{\textbf{Text-Prototype Inference}}} & Normal weather (n=7,811) & $42.59 \pm 4.77$ & $\underline{50.84 \pm 2.65}$ & $34.07 \pm 7.09$ & $\mathbf{56.29 \pm 5.64}$ \\
 & Drought (n=2,299) & $18.66 \pm 1.33$ & $\underline{23.42 \pm 4.14}$ & $14.20 \pm 2.15$ & $\mathbf{25.36 \pm 3.51}$ \\
 & Heat wave (n=684) & $16.81 \pm 3.61$ & $\underline{28.14 \pm 3.79}$ & $15.45 \pm 6.20$ & $\mathbf{43.04 \pm 8.05}$ \\
 & Cold wave (n=466) & $\mathbf{25.49 \pm 5.61}$ & $17.98 \pm 0.71$ & $14.83 \pm 0.81$ & $\underline{24.77 \pm 1.52}$ \\
 & Frost (n=422) & $12.46 \pm 1.73$ & $\underline{33.32 \pm 6.93}$ & $17.74 \pm 7.77$ & $\mathbf{48.48 \pm 15.29}$ \\
 & Atmospheric river (n=4,643) & $31.37 \pm 10.18$ & $\mathbf{72.37 \pm 4.42}$ & $12.70 \pm 1.29$ & $\underline{69.19 \pm 4.15}$ \\
 & Cyclone (n=10,262) & $\underline{54.07 \pm 7.14}$ & $\mathbf{61.67 \pm 5.80}$ & $27.12 \pm 4.06$ & $42.07 \pm 8.38$ \\
 & Convective storm (n=16,092) & $25.20 \pm 4.59$ & $\mathbf{65.39 \pm 0.30}$ & $12.11 \pm 8.89$ & $\underline{57.70 \pm 15.64}$ \\
 & High wind (n=1,864) & $30.31 \pm 1.76$ & $\underline{59.29 \pm 4.24}$ & $29.29 \pm 6.24$ & $\mathbf{66.00 \pm 6.36}$ \\
 & Fog (n=330) & $8.55 \pm 2.22$ & $\underline{33.77 \pm 7.33}$ & $6.92 \pm 2.27$ & $\mathbf{35.31 \pm 10.16}$ \\
 & Winter weather (n=3,772) & $65.89 \pm 3.33$ & $\underline{80.70 \pm 3.24}$ & $52.59 \pm 7.47$ & $\mathbf{84.09 \pm 1.83}$ \\
\rowcolor{TPIMacro}
 & \textbf{Macro F1} $\uparrow$ & $30.13 \pm 2.83$ & $\underline{47.90 \pm 2.09}$ & $21.55 \pm 1.48$ & $\mathbf{50.21 \pm 2.95}$ \\
\midrule
\multirow{12}{*}{\rotatebox[origin=c]{90}{\textbf{Linear Probing}}} & Normal weather (n=7,811) & $90.37 \pm 0.13$ & $\underline{91.74 \pm 0.13}$ & $90.69 \pm 0.12$ & $\mathbf{93.00 \pm 0.04}$ \\
 & Drought (n=2,299) & $48.80 \pm 0.46$ & $\underline{52.54 \pm 0.89}$ & $51.41 \pm 1.17$ & $\mathbf{56.70 \pm 0.61}$ \\
 & Heat wave (n=684) & $47.30 \pm 1.35$ & $\underline{61.94 \pm 1.15}$ & $52.84 \pm 1.49$ & $\mathbf{62.41 \pm 0.86}$ \\
 & Cold wave (n=466) & $48.43 \pm 0.79$ & $\underline{49.49 \pm 0.95}$ & $48.98 \pm 1.93$ & $\mathbf{54.59 \pm 1.23}$ \\
 & Frost (n=422) & $75.38 \pm 1.69$ & $\mathbf{83.14 \pm 0.54}$ & $75.40 \pm 2.56$ & $\underline{82.12 \pm 0.96}$ \\
 & Atmospheric river (n=4,643) & $96.46 \pm 0.18$ & $\mathbf{97.92 \pm 0.10}$ & $96.86 \pm 0.14$ & $\underline{97.73 \pm 0.12}$ \\
 & Cyclone (n=10,262) & $98.11 \pm 0.06$ & $\mathbf{98.31 \pm 0.04}$ & $\underline{98.24 \pm 0.04}$ & $\underline{98.24 \pm 0.02}$ \\
 & Convective storm (n=16,092) & $94.07 \pm 0.09$ & $\mathbf{95.08 \pm 0.16}$ & $94.25 \pm 0.03$ & $\underline{95.03 \pm 0.04}$ \\
 & High wind (n=1,864) & $74.86 \pm 0.55$ & $\mathbf{80.84 \pm 0.74}$ & $78.00 \pm 0.22$ & $\underline{80.03 \pm 0.22}$ \\
 & Fog (n=330) & $62.75 \pm 0.96$ & $\mathbf{76.81 \pm 1.63}$ & $62.36 \pm 0.55$ & $\underline{76.70 \pm 0.55}$ \\
 & Winter weather (n=3,772) & $88.95 \pm 0.24$ & $\mathbf{91.98 \pm 0.08}$ & $90.31 \pm 0.15$ & $\underline{91.73 \pm 0.22}$ \\
\rowcolor{LPMacro}
 & \textbf{Macro F1} $\uparrow$ & $75.04 \pm 0.20$ & $\underline{79.98 \pm 0.26}$ & $76.30 \pm 0.54$ & $\mathbf{80.75 \pm 0.14}$ \\
\bottomrule
\end{tabular}
\end{adjustbox}
\end{table}

\subsection{Physical Baselines for Cyclone Intensity Estimation}
\label{app:physical_baselines}

Since mean sea-level pressure and 10-m wind are included in the ERA5 input, we additionally compare MeteoLIP with physical baselines derived directly from these variables. For minimum central pressure, the direct ERA5 baseline takes the minimum mean sea-level pressure within each input patch, while for maximum sustained wind speed, it takes the maximum 10-m wind speed. We also fit a linear regression from each ERA5-derived quantity to its corresponding IBTrACS target using the training set. As shown in Table~\ref{tab:physical_baselines}, the direct ERA5 baselines achieve $R^2$ values of 51.96 for minimum central pressure and 10.78 for maximum sustained wind speed, which increase to 63.78 and 48.02 with linear regression. In comparison, linear probing of Meteo-CLIP achieves $R^2$ values of 76.36 and 71.34, while Meteo-SigLIP achieves 77.36 and 71.85, respectively. These results show that linear probing of the learned meteorological representations achieves better predictive performance than directly extracting or linearly calibrating the corresponding ERA5 variables.

\begin{table}[t]
\centering
\caption{Physical baselines for cyclone intensity estimation.
 \textbf{Bold} and \underline{underlined} values indicate the best and second-best results, respectively.}
\label{tab:physical_baselines}

\begin{adjustbox}{max width=0.72\linewidth}
\begin{tabular}{@{}llccc@{}}
\toprule
\textbf{Variable} &
\textbf{Method} &
$R^2 \uparrow$ &
\textbf{MAE}$\downarrow$ &
\textbf{RMSE}$\downarrow$ \\
\midrule

\multirow{4}{*}{\textbf{Wind}}
& Direct ERA5
& $10.78$
& $14.50$
& $20.83$ \\

& ERA5 Linear Regression
& $48.02$
& $12.27$
& $15.90$ \\

& \cellcolor{TPIMacro} Meteo-CLIP
& \cellcolor{TPIMacro} \underline{$71.34$}
& \cellcolor{TPIMacro} \underline{$8.65$}
& \cellcolor{TPIMacro} \underline{$11.80$} \\

& \cellcolor{TPIMacro} Meteo-SigLIP
& \cellcolor{TPIMacro} $\mathbf{71.85}$
& \cellcolor{TPIMacro} $\mathbf{8.50}$
& \cellcolor{TPIMacro} $\mathbf{11.70}$ \\

\midrule

\multirow{4}{*}{\textbf{Pressure}}
& Direct ERA5
& $51.96$
& $7.61$
& $13.44$ \\

& ERA5 Linear Regression
& $63.78$
& $8.43$
& $11.67$ \\

& \cellcolor{TPIMacro} Meteo-CLIP
& \cellcolor{TPIMacro} \underline{$76.36$}
& \cellcolor{TPIMacro} \underline{$6.41$}
& \cellcolor{TPIMacro} \underline{$9.43$} \\

& \cellcolor{TPIMacro} Meteo-SigLIP
& \cellcolor{TPIMacro} $\mathbf{77.36}$
& \cellcolor{TPIMacro} $\mathbf{6.16}$
& \cellcolor{TPIMacro} $\mathbf{9.23}$ \\

\bottomrule
\end{tabular}
\end{adjustbox}
\end{table}

{

\subsection{Downstream Fine-Tuning}
\label{app:downstream_finetuning}

Using the same Meteo-SigLIP architecture, we fine-tune models initialized from scratch, the SigLIP WebLI checkpoint, and our Meteo-SigLIP pretrained checkpoint on the classification and physical variable estimation tasks. As shown in Table~\ref{tab:finetuning_initialization}, MeteoLIP initialization achieves the best performance, achieving 81.31 F1 and 92.97 Top-1 accuracy for classification, $R^2$ of 76.15 with 7.56 MAE for maximum sustained wind speed, and $R^2$ of 81.24 with 5.24 MAE for minimum central pressure. These results indicate that meteorology-language pretraining provides a stronger initialization for task-specific fine-tuning, further demonstrating the transferability of the learned multimodal representations.

\begin{table}[t]

\centering
\caption{Effect of initialization on downstream fine-tuning using the same Meteo-SigLIP architecture. \textbf{Bold} and \underline{underline} indicate the best and second-best results, respectively.}
\label{tab:finetuning_initialization}
\small
\setlength{\tabcolsep}{5.5pt}
\renewcommand{\arraystretch}{1.12}
\begin{tabular}{lcccccc}
\toprule
& \multicolumn{2}{c}{\textbf{Classification}} & \multicolumn{2}{c}{\textbf{Sustained Wind}} & \multicolumn{2}{c}{\textbf{Central Pressure}} \\
\cmidrule(lr){2-3}\cmidrule(lr){4-5}\cmidrule(lr){6-7}
\textbf{Initialization} & \textbf{F1} $\uparrow$ & \textbf{Top-1 Acc.} $\uparrow$ & $R^2$ $\uparrow$ & \textbf{MAE} $\downarrow$ & $R^2$ $\uparrow$ & \textbf{MAE} $\downarrow$ \\
\midrule
Scratch & 77.02 $\pm$ 0.76 & 91.56 $\pm$ 0.16 & 72.55 $\pm$ 1.04 & 8.23 $\pm$ 0.18 & 78.89 $\pm$ 0.74 & 5.65 $\pm$ 0.13 \\
WebLI & \underline{78.38 $\pm$ 0.64} & \underline{91.88 $\pm$ 0.20} & \underline{74.31 $\pm$ 0.28} & \underline{7.89 $\pm$ 0.08} & \underline{80.65 $\pm$ 0.37} & \underline{5.36 $\pm$ 0.05} \\
\rowcolor{TPIMacro} MeteoLIP & \textbf{81.31 $\pm$ 0.47} & \textbf{92.97 $\pm$ 0.15} & \textbf{76.15 $\pm$ 0.35} & \textbf{7.56 $\pm$ 0.06} & \textbf{81.24 $\pm$ 0.38} & \textbf{5.24 $\pm$ 0.03} \\
\bottomrule
\end{tabular}
\end{table}
}

{
\subsection{Downstream Transfer to WeatherQA}
\label{app:weatherqa}

We further evaluate the transferability of Meteo-CLIP representations on WeatherQA \citep{weatherqa}, a severe-weather reasoning benchmark over the CONUS, using 7,304 training samples and 600 test samples. WeatherQA comprises a multiple-choice question answering task for the affected geographic region (\textit{Area Affected}) and a classification task for the primary weather concern (\textit{Concerning}).

For each WeatherQA sample, we extract spatial patch representations from the frozen Meteo-CLIP meteorology encoder, map them to the hidden space of Qwen3-VL-4B-Instruct \citep{qwen3vl} using a learnable projection layer, and adapt the language model using LoRA \citep{hu2022lora} with rank 16 and $\alpha=32$. We train for five epochs with AdamW using a batch size of 16, learning rates of $5\times10^{-5}$ and $2\times10^{-4}$ for LoRA and the projection layer, respectively, and a cosine learning-rate schedule.

Meteo-CLIP achieves $49.22 \pm 4.45$\% accuracy on \textit{Area Affected}, which requires selecting the affected region from four candidate options, and $51.73 \pm 2.41$\% conditional accuracy on \textit{Concerning}, an 11-category weather-concern classification task evaluated when the affected area is correctly predicted. These results show that the frozen Meteo-CLIP representations encode spatial and meteorological information that can be effectively transferred to downstream severe-weather reasoning, further demonstrating the usefulness of the learned meteorological representations beyond the EarthAlign evaluation tasks.

\begin{promptbox}{WeatherQA Qualitative Example}
\label{prompt:weatherqa_example}
\footnotesize

\textbf{System Prompt:} \\
You are an AI assistant with expertise in severe weather analysis and forecasting. You are provided with a meteorological representation describing atmospheric and surface conditions across the contiguous United States. Your responsibility is to assess potential severe weather conditions, affected areas, and weather concerns.

\vspace{4pt}
\textbf{User Prompt:}

\texttt{<Parameters>} \\
The provided meteorological representation describes atmospheric and surface conditions across the contiguous United States at 11 / 12, 16UTC.

The meteorological representation is derived from a 63-variable ERA5 snapshot.

\textbf{Pressure-level variables at 300, 500, 700, 850, and 925 hPa:} \\
geopotential, temperature, zonal wind, meridional wind, specific humidity, vertical velocity, and relative vorticity.

\textbf{Surface and single-level variables:} \\
volumetric soil water content in layers 1--4; vertically integrated eastward and northward water-vapour flux; significant wave height; peak wave period; mean wave direction; snowmelt; convective precipitation; snowfall; runoff; total precipitation; convective inhibition; cloud-base height; zero-degree level; instantaneous 10-m wind gust; convective available potential energy; total column water vapour; snow depth; mean sea-level pressure; boundary-layer height; total cloud cover; 10-m zonal and meridional wind; 2-m temperature; and 2-m dewpoint temperature.

\texttt{<meteorological\_representation>} \\
\texttt{</Parameters>}

\vspace{3pt}
\texttt{<Thought process>} \\
Use the provided meteorological representation to answer the following multiple-choice and classification questions:
\begin{enumerate}[leftmargin=*, itemsep=1pt, topsep=2pt, parsep=0pt, partopsep=0pt]
    \item Analyze the spatial and meteorological conditions represented in \texttt{<Parameters>} and identify regions showing signals associated with severe or hazardous weather.
    \item Combine the relevant meteorological information to assess the location and severity of the weather conditions.
    \item Map the conclusions to the provided options. Only give the correct option characters, without any other details or explanations. Use JSON format with the keys \texttt{'Area\_Affected'} for the first answer and \texttt{'Concerning'} for the second answer.
\end{enumerate}
\texttt{</Thought process>}

\vspace{3pt}
\texttt{<Item number="1">}

\textbf{Question:} Choose one geographical area(s) most likely to be impacted by the severe weather event from the four options provided.

\begin{itemize}[leftmargin=*, itemsep=1pt, topsep=2pt, parsep=0pt, partopsep=0pt]
    \item[\textbf{A}] southern/eastern North Carolina and northeastern South Carolina
    \item[\textbf{B}] Central Kansas and Oklahoma
    \item[\textbf{C}] portions of northeast Iowa into far southwest Wisconsin
    \item[\textbf{D}] Far southeast Nebraska...northeast Kansas...extreme northwest Missouri
\end{itemize}

Answer: \texttt{<answer>Option</answer>}

\texttt{</Item>}

\vspace{3pt}
\texttt{<Item number="2">}

\textbf{Question:} Choose the most likely scenario regarding severe weather concerns from the options provided. Options include whether a convective watch has benn issued, the probability of a future watch, or, in winter, potential weather phenomena related to winter storms.

\begin{itemize}[leftmargin=*, itemsep=1pt, topsep=2pt, parsep=0pt, partopsep=0pt]
    \item[\textbf{A}] Severe potential...Watch unlikely (confidence level on the expectation of a watch is 5 or 20\%)
    \item[\textbf{B}] Severe potential...Watch possible (confidence level on the expectation of a watch is 40 or 60\%)
    \item[\textbf{C}] Severe potential...Watch likely (confidence level on the expectation of a watch is 80 or 95\%)
    \item[\textbf{D}] Heavy snow
    \item[\textbf{E}] Severe potential...tornado watch likely (confidence level on the expectation of a watch is 80 or 95\%)
    \item[\textbf{F}] Severe potential...severe thunderstorm watch likely (confidence level on the expectation of a watch is 80 or 95\%)
    \item[\textbf{G}] Winter mixed precipitation
    \item[\textbf{H}] Freezing rain
    \item[\textbf{I}] Severe potential...watch needed soon (confidence level on the expectation of a watch is 95\%)
    \item[\textbf{J}] Blizzard
    \item[\textbf{K}] Snow squall
\end{itemize}

Answer: \texttt{<answer>Option</answer>}

\texttt{</Item>}

\vspace{4pt}
\textbf{Model Output:} \\
{\color{green!50!black}
\texttt{\{"Area\_Affected": "A", "Concerning": "A"\}}
}

\end{promptbox}
}

\subsection{Training Dynamics}
\label{app:training_dynamics}

\begin{figure}[t]
    \centering

    \begin{subfigure}[t]{\textwidth}
        \centering
        \includegraphics[width=\textwidth]{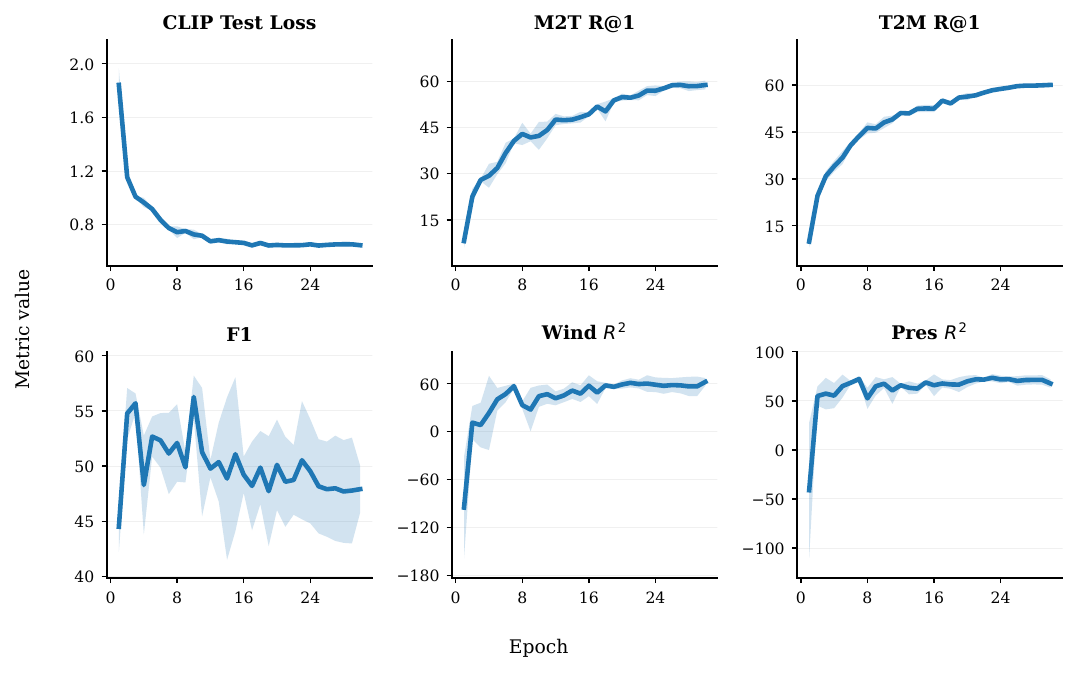}
        \caption{}
        \label{fig:training_dynamics}
    \end{subfigure}

    \vspace{0.5em}

    \begin{subfigure}[t]{\textwidth}
        \centering
        \includegraphics[width=\textwidth]{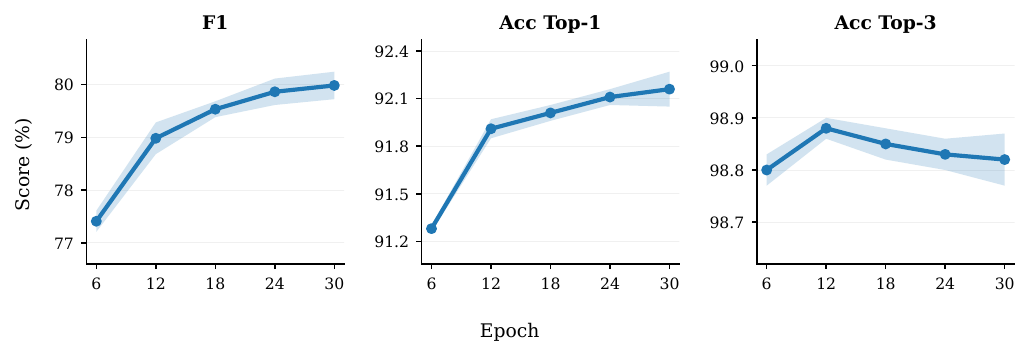}
        \caption{}
        \label{fig:training_dynamics_cls}
    \end{subfigure}

    \caption{Training dynamics of Meteo-CLIP. (a) CLIP {test} loss and retrieval, classification, and physical variable estimation performance across epochs. (b) Linear-probing classification performance using frozen meteorology encoder checkpoints at epochs 6, 12, 18, 24, and 30.}
    \label{fig:training_dynamics_all}
\end{figure}

Figure~\ref{fig:training_dynamics} shows the training dynamics of Meteo-CLIP. As training progresses, the CLIP {test} loss decreases and gradually converges, while retrieval performance generally improves. We observe a similar trend for physical variable estimation, where the $R^2$ values increase substantially after the early training epochs and then remain relatively stable. In contrast, the classification F1 exhibits larger fluctuations and does not show a similarly consistent increasing trend.

To further examine whether these fluctuations reflect the quality of the learned meteorological representations, we perform linear probing on the frozen meteorology encoder checkpoints from epochs 6, 12, 18, 24, and 30. As shown in Figure~\ref{fig:training_dynamics_cls}, both F1 and Top-1 accuracy generally improve with training, while Top-3 accuracy remains stable at approximately 98.8\%. This indicates that although classification performance under text-prototype inference in the learned meteorology-language space can fluctuate across epochs, the frozen meteorology representations continue to become more discriminative for category-level prediction. The nearly saturated Top-3 accuracy further suggests that later training primarily improves the ranking of the correct class among the top candidate categories, which is better reflected by F1 and Top-1 accuracy.

{
\subsection{Positional Encoding in the Meteorological Encoder}
\begin{minipage}[t]{0.59\textwidth}
\vspace{0pt}
We investigate two positional encoding approaches for the meteorological encoder, i.e. learnable positional embeddings (LPE) and 2D rotary positional embeddings (RoPE) \citep{rope2dvit}. For 2D RoPE, we adopt the \emph{RoPE-Mixed} variant, which has been shown to outperform the axial formulation in vision transformers \citep{rope2dvit}. Table~\ref{tab:ablation_lpe_rope} compares LPE, 2D RoPE, and their combination under both OpenAI pre-trained and scratch initialization. LPE achieves the best retrieval performance across both settings, with a larger performance gap when training from scratch. In addition, combining LPE with 2D RoPE does not provide additional gains.
\end{minipage}
\hfill
\begin{minipage}[t]{0.40\textwidth}
\vspace{0pt}
\centering
\captionof{table}{Ablation on positional embedding under pre-trained and scratch initialization.}
\label{tab:ablation_lpe_rope}
\centering
\small
\setlength{\tabcolsep}{4pt}
\renewcommand{\arraystretch}{1.08}
\begin{adjustbox}{max width=\linewidth}
\begin{tabular}{@{}llcc@{}}
\toprule
\textbf{Init.} & \textbf{Pos. Encoding} & \textbf{M2T R@1} $\uparrow$ & \textbf{T2M R@1} $\uparrow$ \\
\midrule
\multirow{3}{*}{OpenAI} & LPE & $\mathbf{58.79 \pm 1.43}$ & $\mathbf{60.10 \pm 0.23}$ \\
 & 2D RoPE & $\underline{54.93 \pm 1.81}$ & $56.29 \pm 1.20$ \\
 & LPE + 2D RoPE & $53.40 \pm 3.19$ & $\underline{56.87 \pm 0.19}$ \\
\midrule
\multirow{3}{*}{Scratch} & LPE & $\mathbf{34.46 \pm 1.78}$ & $\mathbf{41.43 \pm 2.21}$ \\
 & 2D RoPE & $\underline{27.74 \pm 0.69}$ & $\underline{36.03 \pm 0.90}$ \\
 & LPE + 2D RoPE & $25.61 \pm 5.37$ & $35.23 \pm 2.18$ \\
\bottomrule
\end{tabular}
\end{adjustbox}
\end{minipage}
}

\subsection{Model Capacity}
\label{app:model_scaling}

\begin{figure}[t]
    \centering
    \includegraphics[width=\textwidth]{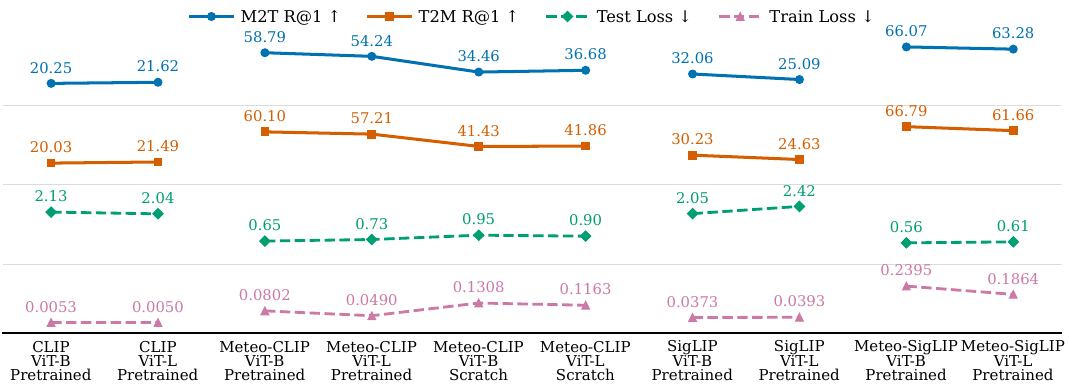}
    \caption{We compare ViT-B and ViT-L encoder variants using pretrained initialization, and additionally compare Meteo-CLIP models trained from scratch. We report M2T and T2M R@1 along with training and test loss to examine how model capacity interacts with initialization and meteorological-language inputs. The reported values are averaged over three training runs.}
    \label{fig:model_scaling}
\end{figure}

Figure~\ref{fig:model_scaling} studies the effect of scaling the encoder from ViT-B to ViT-L. For pretrained Meteo-CLIP, ViT-L achieves a lower training loss than ViT-B (0.0490 vs. 0.0802), but a higher test loss (0.73 vs. 0.65) and lower M2T/T2M R@1 (54.24/57.21 vs. 58.79/60.10). We observe a similar trend for Meteo-SigLIP, where scaling to ViT-L reduces the training loss from 0.2395 to 0.1864, while increasing the test loss from 0.56 to 0.61 and reducing M2T/T2M R@1 from 66.07/66.79 to 63.28/61.66. In comparison, scaling the pretrained CLIP baseline from ViT-B to ViT-L slightly improves both retrieval performance and test loss. When Meteo-CLIP is trained from scratch, ViT-L achieves a lower test loss than ViT-B (0.90 vs. 0.95) and improves M2T/T2M R@1 from 34.46/41.43 to 36.68/41.86. This suggests that although increased capacity can benefit Meteo-CLIP and pretrained visual representations provide a useful initialization (Section~\ref{sec:ablation}), scaling larger pretrained vision-language encoders may require greater adaptation and additional meteorology-language supervision due to the substantial modality gap between RGB images and multivariate meteorological data. These results position \name{} as a challenging setting, where simply fine-tuning a larger model does not guarantee better performance.

{
\section{Limitations and Future Directions}
Although \name{} curates natural event and disaster records from established expert-curated databases and published datasets, their underlying reporting and coverage patterns inevitably shape the geographic, temporal, and event-distribution biases of the dataset. In particular, the United States, Indonesia, and France contain more records than many other countries due to the extensive country-level coverage provided by the NOAA Storm Events Database, IndDE, and FraDE, respectively (Table~\ref{tab:expert_event_sources}). In addition, MeteoLIP currently represents each sample using a single meteorological snapshot. Future work can extend this setting to multiple timestamps and temporally aggregated inputs, as well as broader Earth-system data products such as operational forecasts and other reanalysis datasets. Incorporating regional or downscaled products with higher spatial resolution may further improve the representation of localized phenomena and enable more comprehensive alignment across temporal scales, spatial resolutions, and Earth-system data sources.}

\end{document}